\documentclass[twosides]{article}

\usepackage[accepted]{aistats2021}
\usepackage[round]{natbib}

\usepackage[utf8]{inputenc} 
\usepackage[T1]{fontenc}    
\usepackage{hyperref}       
\usepackage{url}            
\usepackage{booktabs}       
\usepackage{amsfonts}       
\usepackage{nicefrac}       
\usepackage{microtype}      
\usepackage{enumitem}
\usepackage{amsmath}
\usepackage{amsthm}
\usepackage{stackengine}
\usepackage{float}
\usepackage[dvipsnames]{xcolor}
\usepackage{graphicx}
\usepackage{caption}

\newtheorem{assumption}{Assumption}[section]
\newtheorem{theorem}{Theorem}[section]
\newtheorem{corollary}{Corollary}[theorem]
\newtheorem{lemma}[theorem]{Lemma}
\usepackage{array}
\newcolumntype{L}[1]{>{\raggedright\let\newline\\\arraybackslash\hspace{0pt}}m{#1}}
\newcolumntype{C}[1]{>{\centering\let\newline\\\arraybackslash\hspace{0pt}}m{#1}}
\newcolumntype{R}[1]{>{\raggedleft\let\newline\\\arraybackslash\hspace{0pt}}m{#1}}
\makeatletter
\g@addto@macro\normalsize{%
  \setlength\abovedisplayskip{5pt}
  \setlength\belowdisplayskip{5pt}
  \setlength\abovedisplayshortskip{5pt}
  \setlength\belowdisplayshortskip{5pt}
}
\makeatother

\def\TechReport{1}

\begin{document}

%

%

\twocolumn[

\aistatstitle{Towards Flexible Device Participation in Federated Learning}

\aistatsauthor{ Yichen Ruan \And Xiaoxi Zhang \And  Shu-Che Liang \And Carlee Joe-Wong }

\aistatsaddress{ Carnegie Mellon University \And  Sun Yat-Sen University \And Carnegie Mellon University \And Carnegie Mellon University } ]

\begin{abstract}
  Traditional federated learning algorithms impose strict requirements on the participation rates of devices, which limit the potential reach of federated learning. This paper extends the current learning paradigm to include devices that may become inactive, compute incomplete updates, and depart or arrive in the middle of training. We derive analytical results to illustrate how allowing more flexible device participation can affect the learning convergence when data is not independently and identically distributed (non-IID). We then propose a new federated aggregation scheme that converges even when devices may be inactive or return incomplete updates. We also study how the learning process can adapt to early departures or late arrivals, and analyze their impacts on the convergence.
\end{abstract}

\section{Introduction} \label{sec:introduction}
Federated learning is a cutting-edge learning framework that allows distributed devices to train a shared machine learning model cooperatively without sharing the raw data. In recent years, federated learning has exhibited remarkable performance in many applications such as next word suggestion, fault detection, and learning on private medical data \citep{federated_application}. Generic federated learning involves a coordinator and a collection of devices. The training procedure consists of multiple rounds, each of which includes the following three steps: 1) \emph{Synchronization}: the coordinator synchronizes the latest \emph{global model} with all devices. 2) \emph{Local updates}: each device trains a \emph{local model} for a few \emph{local epochs}, using samples from its \emph{local dataset}. 3) \emph{Aggregation}: the coordinator aggregates some, or all, of the local models to produce the next global model.

Our work focuses on cross-device federated learning \citep{advances}, where participating entities are mostly mobile devices such as smart phones and tablets. These devices generally have limited computing and communication resources, e.g., due to battery limitations, and have different training data distributions, i.e., data is not independently and identically distributed (non-IID) among devices \citep{noniid}. To relieve the computation and communication burden, in the last step of the training procedure, the federated learning coordinator may only aggregate a subset of local models. However, only a few device selection policies ensure convergence in the non-IID setting, and the selection must be independent of the hardware status of devices \citep{noniid}. In other words, for the training to converge successfully, all selected devices must be able to train their local models and upload the results whenever they are selected. This is why the traditional federated learning paradigm requires participating devices to be dedicated to the training during the entire federated learning period, e.g., the popular \emph{FedAvg} algorithm assumes mobile users will participate only when their phones are currently plugged-in, and have unlimited WI-FI access \citep{fedavg}. 

Considering that federated learning typically takes thousands of communication rounds to converge, it is difficult to ensure that all devices will be available during the entire training in practice. Moreover, there are typically multiple apps running simultaneously on user devices, competing for already highly constrained hardware resources. As such, it cannot be guaranteed that devices will complete their assigned training tasks in every training round as expected. 
A similar challenge also arises in cloud based distributed learning due to the increasingly popular usage of preemptive cloud services, where the user process can be interrupted unexpectedly \citep{volatile}.

While many methods have been proposed to mitigate the workload of individual devices, such as weight compression and federated dropout \citep{expand}\citep{konevcny2016federated}, they cannot completely remove the possibility that devices are unable to fulfill their training responsibilities, e.g., due to poor wireless connectivity. Thus, in large scale federated learning, many resource-constrained devices have to be excluded from joining federated learning in the first place, which restricts the potential availability of training datasets, and weakens the applicability of federated learning. Furthermore, existing work does not specify how to react when confronting unexpected device behaviors, and also does not analyze the (negative) effects of such behaviors on the training progress. 

In this paper, we relax these restrictions and allow devices to follow more flexible participation patterns. Specifically, the paper incorporates four situations that are not yet well discussed in the literature: 1) \emph{In-completeness}: devices might submit only partially completed work in a round. 2) \emph{Inactivity}: furthermore, devices might not complete any updates, or respond to the coordinator at all. 3) \emph{Early departures}: in the extreme case, existing devices might quit the training without finishing all training rounds.  4) \emph{Late arrivals}: apart from existing devices, new devices might join after the training has already started.

The difference between inactivity and departure is that inactive devices will temporarily disconnect with the coordinator, but are expected to come back in the near future. In contrast, departing devices will inform the coordinator that they do not plan to rejoin the training. For example, if a user quits the app running federated learning, a message can be sent to the coordinator; the coordinator thus knows who is departing. In the meanwhile, although devices' arriving and departing seem symmetric, they affect the model training differently, and thus require distinct treatments. The key difference is that arriving devices offer extra information about the data distribution, which can be utilized to accelerate the training, while departing devices reduce our available knowledge, thus degrading the applicability of the trained model.

Our approach to improve the flexibility of device participation comprises the following components that supplement the existing \emph{FedAvg} algorithm and handle the challenges brought by flexible device participation.
\vspace{-2em}
\begin{itemize} [leftmargin=*]
    \setlength\itemsep{-0.3em}
    \item {\bf Debiasing for partial model updates}. \emph{FedAvg} aggregates device updates as a weighted sum, with weights that are proportional to the sizes of the local datasets. This choice of aggregation coefficients yields an unbiased gradient as in the centralized setting only when all data points from all devices are equally likely to join the learning \citep{noniid}. However, it in general fails to guarantee convergence to the globally optimal point in the presence of partial aggregation from incomplete and inactive devices. We show that by \emph{adapting the aggregation coefficients}, the bias can be reduced and the convergence to a global optimum can still be established. Furthermore, our analysis shows the bias originates from the heterogeneity in device participation, as well as from the degree to which local datasets are not IID.
    \item {\bf Fast-rebooting for device arrivals}.
    Arriving devices interrupt the training by forcing the model to re-orient to the new device's data, thus slowing the convergence process. In this paper, we propose to rapidly reboot the training in the case of device arrivals by applying \emph{extra updates} from the new devices. Intuitively, since an arriving device misses all previous epochs, the model training should emphasize more on its updates to compensate. We will rigorously prove this method indeed expedites learning convergence under certain conditions.
    \item {\bf Redefining model applicability for device departures}. A model successfully trained by federated learning is expected to be applicable to the data from all participating devices. However, when a device withdraws itself from the learning, due to the lack of its future updates, we may no longer require the trained model to perform well on its data. It is then important to redefine the model's applicability. Namely, one can either keep the departing device as a part of the global learning objective, or exclude it to focus only on the remaining devices. The decision depends on which definition yields smaller training loss. We will show the key to this determination lies in the \emph{remaining training time}.
\end{itemize}
\vspace{-0.5em}

In Section \ref{sec:review}, we review relevant literature. In Section \ref{sec:analysis}, we give a convergence analysis that incorporates flexible device participation. Based on this analysis, we detail our contributions, as outlined above, in Section \ref{sec:discussion}, and we experimentally verify our theoretical results in Section \ref{sec:experiment}. Finally we conclude in Section \ref{sec:conclusion}.

\section{Related Works} \label{sec:review}
The celebrated federated learning algorithm named \emph{FedAvg} runs the stochastic gradient descent (SGD) algorithm in parallel on each device in the system and periodically averages the updated parameters from a small set of end devices. However, its performance degrades when the local data is non-IID \citep{hsieh2019non}\citep{weightdivergence}. A few recent works provide theoretical results for the non-IID data case. For instance, \citet{noniid} analyze the convergence of \emph{FedAvg} on non-IID data and establish an $O(\frac{1}{T})$ convergence rate for strongly convex and smooth optimization problems, where $T$ is the number of rounds of local SGD updates. These works either simplify the heterogeneity of the devices, e.g., ignoring cases where some devices may partially finish some aggregation rounds or quit forever during the training~\citep{noniid}, or consider alternative objective functions for the SGD algorithm to optimize~\citep{li2018federated}. Alternatively, some recent papers propose to combine federated learning with the multi-task learning paradigm \citep{smith2017federated}\citep{corinzia2019variational} where multiple models are trained simultaneously, but they also entail dedicated device participation throughout the training.

The \emph{FedAvg} algorithm with non-IID data across devices has also been modified in specific edge computing scenarios to reduce the communication overhead \citep{HierFAVG}\citep{sattler2019robust}\citep{bonawitz2019towards} or maintain a good training convergence under a resource budget constraint \citep{resourcebudget}. However, these works do not consider the possibility that the edge devices can be unavailable during the training process or join at different times, which are the main challenges of this work. An online learning framework \citep{asynchronous}\citep{han2020adaptive}\citep{damaskinos2020fleet} is a possible way to enable flexible device participation in the federated learning scenario. For instance, \citet{asynchronous} propose an asynchronous federated learning algorithm to handle unbalanced data that arrives in an online fashion onto different devices. Although the asynchronous aggregation in their proposed algorithm can be naturally applied to randomly inactive devices, the authors do not analyze how their algorithm's convergence is affected by the device inactivity or incompleteness and the data heterogeneity.

In recent years, some attempts have been made to relax the strict training requirements on the participating devices. For example, \citet{tu2020network} study federated learning in a fog network topology with possible data sharing among devices; \citet{yang2020heterogeneity} incorporate heterogeneity of devices into the design of the learning systems; \citet{nishio2019client} propose a client selection policy that adapts to the change of devices' hardware status. However, these works do not show how the variations in the devices could affect the convergence of training, nor do they incorporate the heterogeneity of user data into the algorithm design.

In \citep{rizk2020dynamic} and \citep{wang2020tackling}, the authors reveal that incomplete devices can block the convergence, but they consider neither other dynamic participation patterns such as inactivity, arrivals and departures, nor probabilistic models for uncertain device participation. To relieve the impact of incomplete devices, these works propose similar strategies as our paper by reweighting the contribution of local models. However, they focus mostly on removing the additional bias term originating from heterogeneous device updates, without looking into how this bias is related to the participation frequency of devices and the divergence among them. They also do not compare the proposed methods with alternative extensions of \emph{FedAvg}. In this work, we model the device participation as random variables and incorporate them into the convergence analysis, and we compare the convergence rates for three reasonable aggregation schemes.
\section{Convergence Analysis} \label{sec:analysis}
In this section, we establish a convergence bound for federated learning with flexible device participation patterns. Our analysis generalizes the standard \emph{FedAvg} to incorporate arbitrary aggregation coefficients. In the aggregation step, all devices are counted even if they cannot finish all local epochs. The analysis considers a non-IID data distribution and heterogeneous devices, i.e., some devices can be more stable than the others. We first derive the convergence bound with incomplete and inactive devices in Sections \ref{ssec:algorithm} to \ref{ssec:bound}, and then discuss arrivals and departures in Section \ref{ssec:shift}.  

\subsection{Algorithm Description} \label{ssec:algorithm}
Suppose there are $N$ devices, where each device $k$ defines a local objective function $F_k(w)$. Here $w$ represents the parameters of the machine learning model to be optimized, and $F_k(w)$ may be defined as the average empirical loss over all data points at device $k$, as in typical federated learning frameworks \citep{fedavg}. The global objective is to minimize $F(w) = \sum_{k=1}^N p^k F_k(w)$, where $p^k = \frac{n_k}{n}$, $n_k$ is the number of data points device $k$ owns, and $n = \sum_{k=1}^N n_k$. Let $w^*$ be the minimizer of $F$, and denote by $F_k^*$ the minimum value of $F_k$. We quantify the degree to which data at each device $k$ is distributed differently than that at other devices as $\Gamma_k = F_k(w^*) - F_k^*$ to capture that data distributions at different devices are non-IID, and let $\Gamma = \sum_{k=1}^N p^k\Gamma_k$ as in \citep{noniid}.

We consider discrete time steps $t=0,1,\dots$. Model weights are synchronized when $t$ is a multiple of $E$, i.e., each round consists of E time steps. Assume there are at most $T$ rounds. For each round (say the $\tau$th round), the following three steps are executed:

\vspace{-0.5em}
\begin{itemize}[leftmargin=*]
    \setlength\itemsep{-0.3em}
    \item Synchronization: the coordinator broadcasts the latest global weight $w_{\tau E}^{\mathcal{G}}$ to all devices. Each device updates its local weight so that: $w_{\tau E}^k = w_{\tau E}^{\mathcal{G}}$
    \item Local updates: each device runs stochastic gradient descent (SGD) on $F_k$ for $i=0,\dots,s_\tau^k-1$:\footnote{While some papers define local epochs and local updates separately, we use them interchangeably in this paper. Both refer to the times (\ref{eq:local_sgd_raw}) is conducted in a global round.}
    \begin{equation} \label{eq:local_sgd_raw}
        w_{\tau E + i + 1}^k = w_{\tau E + i}^k - \eta_{\tau} g_{\tau E +i}^k
    \end{equation}
    Here $\eta_\tau$ is a staircase learning rate that decays with $\tau$, $0 \leq s_\tau^k \leq E$ represents the number of local updates this device completes in this round, $g_{t}^k = \nabla F_k (w_{t}^k, \xi_{t}^k)$ is the stochastic gradient at device $k$, and $\xi_{t}^k$ is a mini-batch sampled from device $k$'s local dataset. We also define $\bar{g}_{t}^k = \nabla F_k(w_t^k)$ as the full batch gradient at device $k$, hence $\bar{g}_t^k = \mathbb{E}_{\xi_t^k}[g_t^k]$. 
    \item Aggregation: the coordinator aggregates the gradients and generates the next global weight as
    \begin{equation} \label{eq:aggregation_raw}
    \begin{split}
        w_{(\tau+1)E}^\mathcal{G} &= w_{\tau E}^\mathcal{G} + \sum\nolimits_{k=1}^N p_\tau^k (w_{\tau E + s_\tau^k} - w_{\tau E}^\mathcal{G})\\
        &= w_{\tau E}^\mathcal{G} - \sum\nolimits_{k=1}^N p_\tau^k \sum\nolimits_{i=0}^{s_\tau^k} \eta_{\tau}g_{\tau E + i}^k
    \end{split}
    \end{equation}
\end{itemize}
\vspace{-0.5em}

We define that a device $k$ is \emph{inactive} in round $\tau$ if $s_\tau^k = 0$ (i.e., it completes no local updates), and say it is \emph{incomplete} if $0 < s_\tau^k < E$. We treat each $s_\tau^k$ as a random variable that can follow an arbitrary distribution. Devices are \emph{heterogeneous} if they have different distributions of $s_\tau^k$, and otherwise they are \emph{homogeneous}. We allow the aggregation coefficients $p_\tau^k$ to vary with $\tau$. In Section \ref{sec:discussion}, we will discuss different schemes of choosing $p_\tau^k$ and their impacts on the convergence.
    
As a special case, traditional \emph{FedAvg} assumes all selected devices can complete all $E$ local epochs, so that $s_\tau^k \equiv E$. Also, \emph{FedAvg} with full device participation uses fixed aggregation coefficients $p_\tau^k \equiv p^k$, so that the right hand side of (\ref{eq:aggregation_raw}) can be written as $\sum_{k=1}^N p^k w_{\tau E}^k$, i.e., aggregating gradients is equivalent to aggregating the model parameters directly.


\subsection{General Convergence Bound} \label{ssec:bound}
The analysis relies on the following five assumptions. The first four are standard \citep{noniid}. The last assumption ensures bounded aggregation coefficients and is satisfied by all schemes discussed in Section \ref{sec:discussion}. In Section \ref{sec:experiment}, we experimentally show that our proposed learning algorithm performs well even when some assumptions (like strong convexity) are violated.

\begin{assumption} \label{asm:Lsmooth}
$F_1,\dots,F_N$ are all $L$-smooth, so that $F$ is also $L$-smooth.
\end{assumption}
\begin{assumption} \label{asm:mu_strong}
$F_1,\dots,F_N$ are all $\mu$-strongly convex, so that $F$ is also $\mu$-strongly convex.
\end{assumption}
\begin{assumption} \label{asm:sigma2}
The variance of the stochastic gradients is bounded: $\mathbb{E}_{\xi} \Vert g_t^k - \bar{g}_t^k \Vert^2 \leq \sigma_k^2$, $\forall k,t$.
\end{assumption}
\begin{assumption} \label{asm:G2}
The expected squared norm of the stochastic gradients at each local device is uniformly bounded: $\mathbb{E}_\xi \Vert g_t^k \Vert^2 \leq G^2$ for all $k$ and $t$.
\end{assumption}
\begin{assumption} \label{asm:cp}
There exists an upper bound $\theta > 0$ for the aggregation coefficient: $p_\tau^k/p^k \leq \theta, \forall k$.
\end{assumption}

Assume the following expectations exist and do not vary with time: $\mathbb{E}[p_\tau^k]$, $\mathbb{E}[p_\tau^ks_\tau^k]$, $\mathbb{E}[(p_\tau^k)^2s_\tau^k]$, $\mathbb{E}[(\sum_{k=1}^Np_\tau^k-2)_+(\sum_{k=1}^Np_\tau^ks_\tau^k)]$ for all rounds $\tau$ and devices $k$, and assume $\mathbb{E}[\sum_{k=1}^N p_\tau^ks_\tau^k] \neq 0$. Intuitively, this last assumption ensures that some updates are aggregated in each round, otherwise this round can be simply omitted. Generally, $p_\tau^k$'s are functions of $s_\tau^k$, and these expectations can be estimated from device histories. Let $z_\tau \in \{0,1\}$ indicate the event that the ratio $\mathbb{E}[p_\tau^ks_\tau^k]/p^k$ does not take the same value for all $k$. We can obtain the following convergence bound for general $p_\tau^k$:



\begin{theorem} \label{tm:convergence0}

By choosing the learning rate $\eta_\tau = \frac{16E}{\mu \mathbb{E}[\sum_{k=1}^Np_\tau^ks_\tau^k]}\frac{1}{\tau E + \gamma}$, we can obtain
\begin{equation} \label{eq:convergence0}
    \mathbb{E}\|w_{\tau E}^{\mathcal{G}} - w^* \|^2 \leq \frac{M_\tau D + V}{\tau E + \gamma}
\end{equation}

Here we define $\gamma = \max \left\{\frac{32E(1+\theta)L}{\mu \mathbb{E}[\sum_{k=1}^Np_\tau^ks_\tau^k]}, \frac{4E^2\theta}{ \mathbb{E}[\sum_{k=1}^Np_\tau^ks_\tau^k]} \right\}$, $M_\tau = \sum_{t=0}^{\tau - 1} \mathbb{E}[z_t]$, $D=\frac{64E\sum_{k=1}^N\mathbb{E}[p_\tau^ks_\tau^k]\Gamma_k}{\mu \mathbb{E}[\sum_{k=1}^Np_\tau^ks_\tau^k]}$, $V = \max\left\{\gamma^2 \mathbb{E}\|w_{0}^{\mathcal{G}} - w^* \|^2, \left(\frac{16E}{\mu\mathbb{E}[\sum_{k=1}^Np_\tau^ks_\tau^k]}\right)^2 \frac{\mathbb{E}[B_\tau]}{E} \right\}$, $B_\tau = 2(2+\theta)L \sum_{k=1}^N p_\tau^ks_\tau^k \Gamma_k + \left(2 + \frac{\mu}{2(1+\theta)L}\right)E(E-1)G^2 \left(\sum_{k=1}^Np_\tau^k s_\tau^k + \theta(\sum_{k=1}^N p_\tau^k - 2)_+\sum_{k=1}^N p_\tau^ks_\tau^k \right) + 2EG^2\sum_{k=1}^N\frac{(p_\tau^k)^2}{p^k}s_\tau^k + \sum_{k=1}^N (p_\tau^k)^2s_\tau^k \sigma_k^2$
\end{theorem}

Theorem 3.1 shows that the convergence rate is affected by the aggregation coefficients $p_{\tau}^k$'s as they determine $M_{\tau}$, $D$, and $V$.
From (\ref{eq:convergence0}), $w_{\tau E}^{\mathcal{G}}$ will eventually converge to a globally optimal solution only if $M_\tau$ increases sub-linearly with $\tau$. In the original full-participation \emph{FedAvg}, $p_\tau^ks_\tau^k \equiv p^kE$, so $z_\tau \equiv 0$ and $M_\tau \equiv 0$ as per the definitions. Thus, full-participation \emph{FedAvg} converges according to (\ref{eq:convergence0}), which is consistent with \citep{noniid}. However, when considering flexible device participation, $M_\tau$ may increase with $\tau$, which can cause $\emph{FedAvg}$ to converge to an arbitrary suboptimal point. The magnitude of $M_\tau$ is determined by the degree of heterogeneity in the device participation, and $D$ is bounded by the non-IID metric $\Gamma_k$ of local datasets. If $M_\tau$ increases linearly with $\tau$ (e.g., due to device departures), the model will converge to a suboptimal point with the loss bounded by $\frac{D}{E}$. As we will see in Section \ref{ssec:schemes}, by smartly choosing the aggregation coefficients $p_\tau^k$, the increase of $M_\tau$ can be controlled and a convergence to the global optimum can still be established.

While we only show results for $s_\tau^k$ whose distributions are static with time, Theorem \ref{tm:convergence0} can be easily extended to time-varying distributed $s_\tau^k$ by replacing the corresponding expectations of $p_\tau^ks_\tau^k$ and $(p_\tau^k)^2s_\tau^k$ with their minimum or maximum expectations over $\tau$.

\subsection{Shifts in the Global Objective} \label{ssec:shift}
Recall the global objective is $F(w) = \sum_{k \in \mathcal{C}} p^kF_k(w)$, i.e., an average of local objectives for participating devices $\mathcal{C}$. A well trained model $w^*$ is expected to perform well on all data points generated by devices in $\mathcal{C}$. In the presence of departing and arriving devices, $\mathcal{C}$ may shrink or expand dynamically during the training. The global objective thus varies accordingly. For example, after admitting an incoming device $l$ with $n_l$ data points: $\Tilde{\mathcal{C}} \gets \mathcal{C} + \{l\}$, the global objective becomes $\Tilde{F}(w) = \Tilde{p}^lF_l(w) + \sum\nolimits_{k \in \mathcal{C}} \Tilde{p}^kF_k(w)$, where $\Tilde{p}^k = \frac{n_k}{n_{\mathcal{C}} + n_l}$. The model $\Tilde{w}^*$ fully trained with this objective is then applicable to the new data from device $l$. We formally define \emph{objective shift} as the process of changing the global objective, and the applicability of the trained model, by adding or removing devices from $\mathcal{C}$. 

The following theorem bounds the offset between the global optima due to the objective shift. As we can intuitively expect, the difference reduces when the data becomes more IID ($\Gamma_l \rightarrow 0$), and when the departing/arriving device owns fewer data points ($n_l \rightarrow 0$):

\begin{theorem} \label{tm:shift}
Suppose a device $l$ arrives/departs, and let $n$ be the total number of data points originally. Consider the objective shift $F \rightarrow \Tilde{F}$, $w^* \rightarrow \Tilde{w}^*$. Let $\Tilde{\Gamma}_k = F_k(\Tilde{w}^*) - F_k^*$ quantify the degree of non-IID with respect to the new objective. Then in the arrival case
\begin{equation} \label{eq:shift_arrive}
    \|w^* - \Tilde{w}^* \| \leq \frac{2\sqrt{2L}}{\mu} \frac{n_l}{n+n_l}\sqrt{\Gamma_l}
\end{equation}
and in the departure case
\begin{equation} \label{eq:shift_depart}
    \|w^* - \Tilde{w}^* \| \leq \frac{2\sqrt{2L}}{\mu} \frac{n_l}{n}\sqrt{\Tilde{\Gamma}_l}
\end{equation}

\end{theorem}

Objective shift is mandatory when a new device (say device $l$) arrives: Unless $F_l \equiv F$ (which is highly unlikely), incorporating updates from $l$ will always move $F(w)$ away from $F^*$. The best strategy without objective shift is then not to aggregate updates from $l$, and thus not to admit $l$ into the learning process in the first place. In contrast, objective shift is optional when devices depart: we can keep the original objective $F$ even if we will no longer receive updates from a departing device, if doing so yields smaller training loss. 


Suppose an objective shift occurs at $\tau_0$. The remainder of the training is then equivalent to starting over from $w_{\tau_0 E}^\mathcal{G}$ but converging towards the new objective $\Tilde{w}^*$. Combining Theorems \ref{tm:convergence0} and \ref{tm:shift}, we can obtain the following convergence bound after the objective shifts:



\begin{corollary} \label{coro:convergence_shift}
Assume the objective shifts at $\tau_0$ with $\mathbb{E}\|w_{\tau_0 E}^{\mathcal{G}} - w^* \|^2 \leq \Delta_{\tau_0}$. By increasing the learning rate back to $\eta_{\tau} = \frac{16E}{\mu \mathbb{E}[\sum_{k=1}^Np_\tau^ks_\tau^k]}\frac{1}{(\tau-\tau_0) E + \gamma}$ for $\tau > \tau_0$, the convergence to the new objective can be bounded by
\begin{equation} \label{eq:convergence_shift}
\begin{split}
    \mathbb{E}\|w_{\tau E}^{\mathcal{G}} &- \Tilde{w}^* \|^2 \leq \frac{\Tilde{M}_\tau \Tilde{D} + \Tilde{V}}{(\tau - \tau_0) E + \Tilde{\gamma}}
\end{split}
\end{equation}

Here $\Tilde{M}_\tau, \Tilde{D}, \Tilde{V}, \Tilde{\gamma}$ are defined analogously to $M_\tau$, $D$, $V$, $\gamma$ but they respectively include/exclude the arriving/departing device. The first term in $\Tilde{V}$ equals $\Tilde{\gamma}^2 (\sqrt{\Delta_{\tau_0}} + \|w^* - \Tilde{w}^* \|)^2 = O\left(\frac{V}{\tau_0E + \gamma} + \Gamma_l\right)$.
\end{corollary}

The increase of the learning rate after the objective shift is necessary. Intuitively, if the shift happens at a large time $\tau_0$ when $w_{\tau_0 E}^{\mathcal{G}}$ is close to the old optimal $w^*$ and $\eta_{\tau_0}$ is close to zero, the learning rate used in Theorem \ref{tm:convergence0} will be too small to steer the model to the new optimum, since $\|w_{\tau_0 E}^{\mathcal{G}} - \Tilde{w}^* \| \approx \|w^* - \Tilde{w}^* \|$.

Comparing (\ref{eq:convergence0}) and (\ref{eq:convergence_shift}), an objective shift yields an one-time increase in the loss, which forces us to take actions when confronting departures and arrivals. In the case of device departure, it is possible that retaining the old objective can result in a smaller training loss compared to doing a shift. In this situation, the trained model is still applicable to data of the departing device. In the arrival case, though objective shift is mandatory, we can still accelerate the training by a ``fast-reboot'', applying extra gradient updates from the arriving device.

We will discuss in Section \ref{ssec:fast_reboot} the fast-reboot method for the arrival case, and in Section \ref{ssec:applicability} the decision of model applicability for the departure case.
\section{Main Results} \label{sec:discussion}
Based on the convergence analysis in Section \ref{sec:analysis}, in this section, we present corollaries that can guide operators in reacting to flexible device participation.

\subsection{Debiasing on Incomplete Aggregation} \label{ssec:schemes}

According to Theorem \ref{tm:convergence0}, the convergence bound is controlled by the expectation of $p_\tau^k$ and its functions. Below we discuss three plausible schemes of choosing $p_\tau^k$, and compare their convergence rates in Table \ref{tab:compare}.

\vspace{-1em}
\begin{itemize}[leftmargin=*]
    \setlength\itemsep{-0.3em}
    \item Scheme A: Only aggregate parameters from devices that complete all $E$ local epochs, with aggregation coefficient $p_\tau^k = \frac{Np^k}{K_\tau}q_\tau^k$, where $K_\tau$ is the number of complete devices, $q_\tau^k \in \{0,1\}$ denotes if client $k$ is complete. If $K_\tau=0$, this round is discarded.
    \item Scheme B: Allow clients to upload incomplete work  (with $s^k_\tau < E$ updates), with fixed aggregation coefficient $p_\tau^k = p^k$.
    \item Scheme C: Accept incomplete works as in Scheme B, with adaptive $p_\tau^k =  \frac{E}{s_\tau^k}p^k$, or $p_\tau^k = 0$ if $s_\tau^k = 0$.
\end{itemize}
\vspace{-1em}

Schemes A and B are natural extensions of \emph{FedAvg}. Scheme C assigns a greater aggregation coefficient to devices that complete fewer local epochs. Though this idea seems counter-intuitive, as fewer local updates might lead to less optimal parameters (cf. Table \ref{tab:compare}), it turns out to be the only scheme that guarantees convergence when device participation is heterogeneous.

\begin{corollary} \label{cor:compare}
Let $K_\tau$ be the number of devices that run all $E$ epochs, $I_\tau$ indicate the appearance of any inactive devices in round $\tau$, and write $\Bar{\sigma}^2_N \equiv \sum_{k}^N(p^k\sigma_k)^2$. Table \ref{tab:compare} gives the convergence rates of Schemes A, B, C when device updates may be incomplete and inactive.

\end{corollary}

\begin{table}[h]
\setlength{\tabcolsep}{3pt}
\begin{center}
\caption{\label{tab:compare} Convergence rates with incomplete and inactive devices. The bound for Scheme A assumes there is at least one complete device ($K_\tau \neq 0$), and those for Schemes B, C assume $s_\tau^k$ is not trivially zero ($\mathbb{E}[s_\tau^k] \neq 0$). While the three schemes have similar performance in the homogeneous setting, Schemes A and B fail to converge to the global optimum even assuming all devices are active. Scheme C works if inactive devices do not occur in every round ($\sum_t I_t < O(\tau)$).}
\begin{tabular}{ c| c |c } 
 \hline
          & \textbf{Homogeneous} & \textbf{Heterogeneous} \\
 \hline
  \textbf{A} & $O\left(\frac{\mathbb{E}[\frac{N^2}{K_\tau}]+\Bar{\sigma}^2_N + \Gamma}{\tau}\right)$ & $\leq  \frac{D}{E}$  \\ 
 \hline
  \textbf{B} & $O\left(\frac{ \Bar{\sigma}^2_N + \Gamma}{\tau\mathbb{E}[s_\tau]}\right)$ & $\leq \frac{D}{E}$  \\ 
 \hline
  \textbf{C} & $O\left(\frac{\Bar{\sigma}^2_N+ \Gamma}{\tau(\mathbb{E}\left[1/s_\tau\right])^{-1}}\right)$ & $O\left( \frac{\sum\limits_{t=0}^{\tau - 1}I_t D + \sum\limits_{k}^N(p^k\sigma_k)^2\mathbb{E}\left[\frac{1}{s_\tau^k}\right] + \Gamma}{\tau}\right)$  \\ 
 \hline
\end{tabular}
\end{center}
\end{table}

\begin{figure}[h] 
    \centering
    \includegraphics[width=0.4\textwidth]{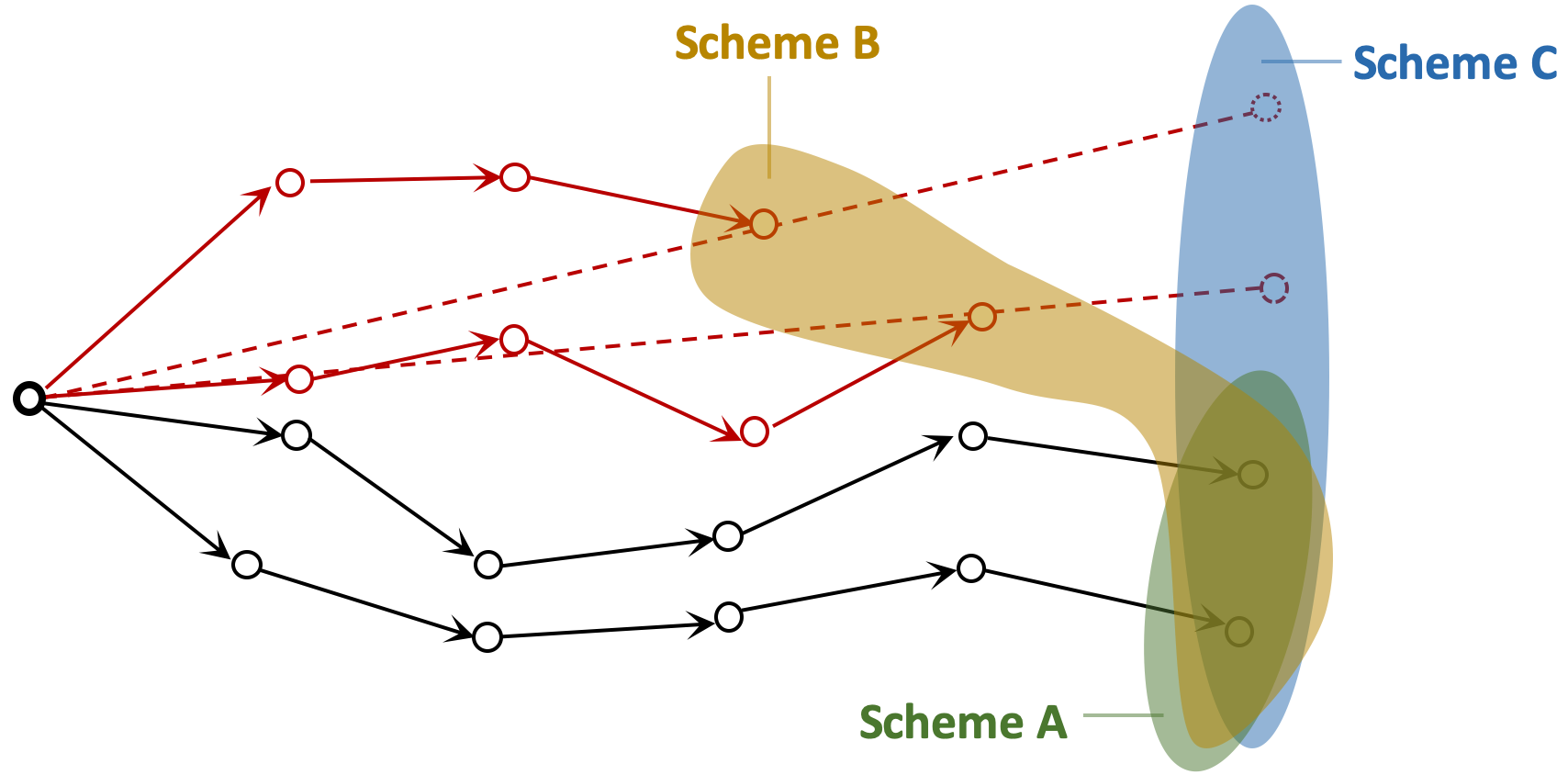}
    \caption{Snapshot of one aggregation round. The bottom two devices completed all $E=5$ local epochs, while the top two completed only 3 and 4 epochs. Scheme C enlarges the incomplete gradients by respectively $5/3$ and $5/4$ and produces unbiased aggregation results. Aggregations with Schemes A and B are biased towards devices that run more epochs.\vspace{-1em}}\label{fig:converge}
\end{figure}

The reason for enlarging the aggregation coefficients in Scheme C can be understood by observing from (\ref{eq:aggregation_raw}) that increasing $p_\tau^k$ is equivalent to increasing the learning rate of device $k$. Thus, by assigning devices that run fewer epochs a greater aggregation coefficient, these devices effectively run further in each local step, compensating for the additional epochs other devices completed. As shown in Figure \ref{fig:converge}, Scheme C ensures an unbiased gradient after aggregation, while Schemes A and B will favor devices that run more epochs. Ideally, allowing devices to adapt learning rates by themselves would effectively lead to the same result. However, when a device is running local updates, it may not yet know or be able to estimate the number of local epochs it will complete. In contrast, centralized intervention can make accurate adjustments a posterior.

Table \ref{tab:compare} also reveals how the following system and statistical factors affect the convergence asymptotically:
\begin{itemize} [leftmargin=*]
    \setlength\itemsep{-0.3em}
    \item The non-IID metric $\Gamma$ is the major obstacle of convergence in the homogeneous case. In the heterogeneous setting, the $D$ term (which grows with $\Gamma$) dominates the training loss. It controls the maximum non-diminishing loss $D/E$ of Scheme A and B, and decelerates the training of Scheme C in the presence of inactive devices.
    \item Devices' activeness $s_\tau^k$ and $K_\tau$ contribute inversely to the training loss: The more devices participate, the faster the loss decays. When inactivity occurs frequently, Scheme C cannot converge either. E.g., if a device never responds to the coordinator (so $I_t \equiv 1$), its training loss can never converge to zero.
    \item The variance $\Bar{\sigma}_N, \sigma_k$ in the stochastic gradient descent algorithm slows down the training as expected.
\end{itemize}
\vspace{-1em}

\subsection{Fast-rebooting on Arrivals} \label{ssec:fast_reboot}

\begin{figure}[h]
    \centering
    \includegraphics[width=0.45\textwidth]{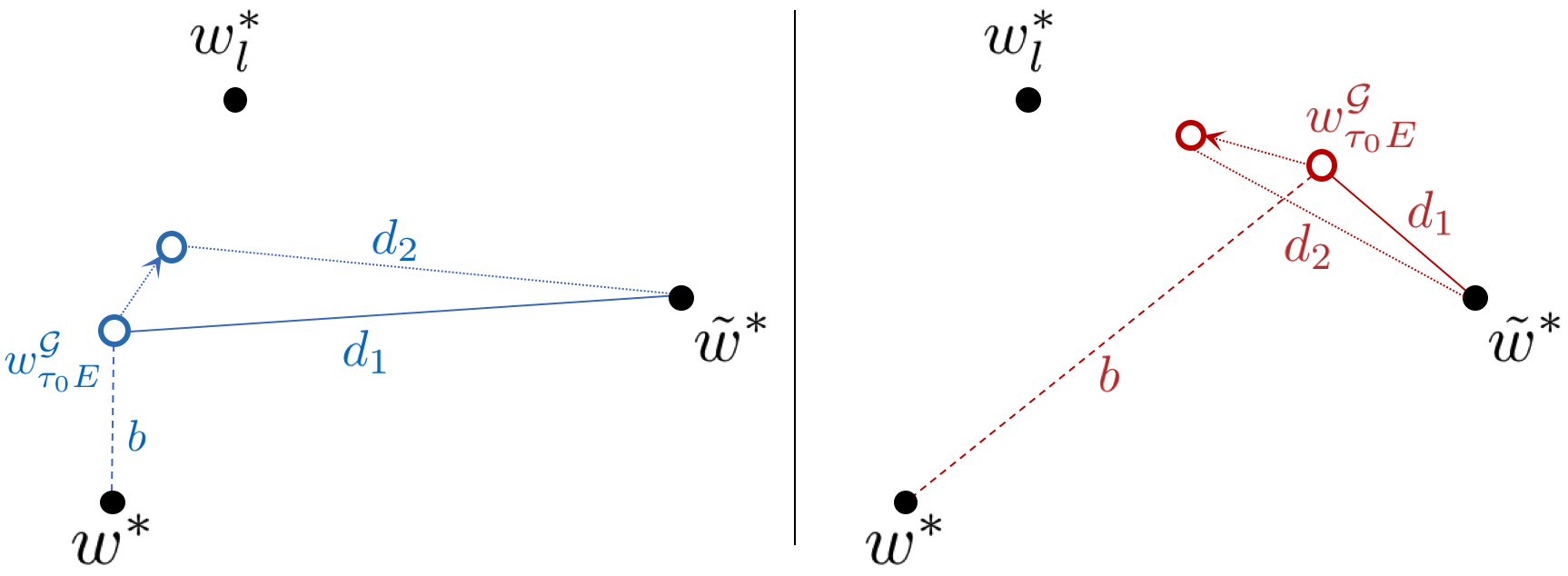}
    \caption{Left: when the distance to the old optimum $b = \|w^{\mathcal{G}} - w^*\|$ is small, applying an extra update to $w^{\mathcal{G}}$ following the direction $-\nabla F_l(w^{\mathcal{G}})$ moves it closer to $\Tilde{w}^*$ ($d_2 < d_1$). Right: for a large $b$, the extra update may on the contrary enlarge this distance ($d_2 > d_1$).}\label{fig:arrival}
\end{figure}

Intuitively, when a device $l$ arrives, $\Tilde{w}^*$ will be ``dragged'' towards its local optimum $w_l^*$. The gradients from device $l$ may thus encode more information about the new optimum $\Tilde{w}^*$ compared to those from the other devices. Thus, by adding an extra update $-\delta^l \nabla F_l(w^{\mathcal{G}}), \delta^l > 0$ to the gradient aggregation, it is likely that $w$ can move closer to $\Tilde{w}^*$, allowing the training to fast-reboot from the point of arrivals. However, as shown in Figure \ref{fig:arrival}, this intuition may not hold: it is also possible that $-\delta^l \nabla F_l(w^{\mathcal{G}})$ ends up driving $w^{\mathcal{G}}$ away from $\Tilde{w}^*$. In fact, the success of this method is determined by the distance $b = \|w^{\mathcal{G}} - w^*\|$. When $b$ is small, it is highly likely the extra update can rapidly reboot the training. We formalize this statement in Corollary \ref{coro:fast_reboot}.

\begin{corollary} \label{coro:fast_reboot}
Assume $\nabla F(w)$ is continuous, and $0 < \|\nabla F(w) \|_2, \|\nabla^2 F(w)\|_2 \leq W$ for any $w$ (The latter is the induced $l_2$ norm for matrices). Let $w' = w - \delta^l \nabla F_l(w)$, then there exists a $\delta^l > 0$ such that $\|w'-\Tilde{w}^* \| < \|w - \Tilde{w}^*\|$ if $w$ satisfies
\begin{equation} \label{eq:fast_reboot}
    \|w - w^* \| < \frac{\Tilde{F}(w^*)-\Tilde{F}(\Tilde{w}^*)}{\left(\frac{2\sqrt{2L}}{\mu} \Tilde{p}^l\sqrt{\Gamma_l} + 1\right)\Tilde{p}^lW}
\end{equation}
\end{corollary}

(\ref{eq:fast_reboot}) defines a sphere around the original global optimum $w^*$ within which the extra update helps fast-reboot. The radius of the sphere depends on the divergence between the new (arriving) and old data points. Generally, the longer the training has elapsed, the closer the global model is to $w^*$. Thus, the extra updating works best for devices that arrive late in the training. 

When applied in practice, the extra updating can be conducted on-the-fly, by augmenting the aggregation coefficient of the arriving device so that $p_\tau^l = p^l + \delta^l$. Furthermore, the distance $b$ can be estimated by the gradient norm with respect to the original objective.

As the name suggests, fast-reboot only accelerates the training for a certain duration after the device arrives. In fact, if there are no future interrupts, models with or without fast-rebooting eventually converge to the same global optimum. Nevertheless, fast-reboot is still beneficial if there is insufficient training time remaining (e.g., a device arrives near the end of the training).

\subsection{Redefining Applicability on Departures} \label{ssec:applicability}
As is discussed in Section \ref{ssec:shift}, when a device leaves, we need to redefine the applicability of the trained model. Namely, one can decide to either exclude this departing device and shift the objective, or keep including it and stick to the old objective. The decision depends on the time at which the device leaves. When including the device as a part of the global objective, from (\ref{eq:convergence0}), since $M_\tau = \tau - \tau_0$ from then on, the training loss will always exceed a structural bias $D/E$. In contrast, if the device is excluded and the model is trained with a shifted global objective, there will be an immediate increase in the convergence bound as in Theorem \ref{tm:shift}. But afterwards, the bound will decrease and eventually the parameters will converge to the new global optimum.


Assume a device leaves at $\tau_0 < T$ and there are no subsequent arrivals/departures. Let $f_0(\tau)$ be the convergence bound if we include the device, and $f_1(\tau)$ be the bound if it is excluded. We can obtain $f_{\textrm{0}}(\tau) = \frac{(\tau - \tau_0)D + V}{\tau E + \gamma}, f_1(\tau) = \frac{\Tilde{V}}{(\tau - \tau_0)E + \Tilde{\gamma}} $. Here $\Tilde{M}_\tau, \Tilde{V}_\tau, \Tilde{\gamma}$ are defined analogously to $M_\tau, V_\tau, \gamma$ but they exclude the departing device. A device is excluded if by doing so, a smaller training loss can be obtained at the deadline $T$, which is summarized in the following corollary:

\begin{corollary} \label{cor:departure}
Excluding a device that departs at $\tau_0$ leads to smaller training loss if
\begin{equation} \label{eq:depature_decision}
     \min_{\tau \geq \tau_0} f_0(\tau)
     \geq f_1(T)
\end{equation}
Further assume $\Tilde{\gamma} = \gamma$, and $\Tilde{V}$ is dominated by its first term so that $\Tilde{V} = \frac{V}{\tau_0 E + \gamma} + \Gamma_l$. (\ref{eq:depature_decision}) then becomes
\begin{equation} \label{eq:threshold_tau}
     T - \tau_0
     \geq O\left(\sqrt{\Gamma_l\tau_0}\right)
\end{equation}
\end{corollary}

From (\ref{eq:threshold_tau}), when the remaining training time $T - \tau_0$ is at least $O(\sqrt{\Gamma_l\tau_0})$, applying the trained model to the departing device becomes less promising. It is thus better to exclude it and shift the objective. As we can expect, the bound grows with $\Gamma_l$, since the non-IID contribution from the departing device increases the initial $\Tilde{V}$. As $\tau_0$ increases, the learning rate without shift gets smaller, mitigating the increase of the training loss from departing devices.
\section{Experiments} \label{sec:experiment}
In this section, we experimentally evaluate Section \ref{sec:discussion}'s results. Due to the limitations on hardware resources, the training process is performed in computer simulations. To ensure the simulation is consistent with the real learning environment, we use real-world traces to represent the participation patterns of simulated devices. We present our experiment setup in Section \ref{ssec:exp_setup}, and verify our theory results in Sections \ref{ssec:exp_schemes} - \ref{ssec:exp_departure}.

\subsection{Experiment Setup} \label{ssec:exp_setup}

We create various data traces to represent the heterogeneous participation patterns of local devices. We set up a simple federated learning experiment with five Raspberry PIs as workers, and a desktop server as the coordinator. Each PI has a training process that runs the original \emph{FedAvg} algorithm, and a competitor process doing CPU-intensive work simultaneously. We manually tune the workload of the competitor process so that it takes up 0\%, 30\%, 50\%, 70\%, 90\% of the PI's CPU resources, simulating different device configurations in federated learning. Under the five settings, for each round, we record the percentage of required epochs the PI ends up submitting before a preset, fixed deadline. Due to the default load-balancing behavior of the operating system's CPU scheduler, these traces do not contain zero epochs (i.e. inactive cases). To generate inactive device participation patterns, we create another set of three traces with respectively low, medium and high bandwidth. Devices can thus be inactive due to weak transmission. Table \ref{tab:traces} shows the mean and standard deviation of the percentage of epochs completed for each trace. In the following experiments, each simulated device is randomly assigned a trace. For each aggregation round $\tau$, it randomly samples from its trace to obtain the number of local epochs $s_\tau^k$.

\begin{table}[h]
\setlength{\tabcolsep}{3pt}
\begin{center}
\caption{\label{tab:traces} The means and standard deviations for the percentage of required local epochs actually submitted to the coordinator during the federated training. The first five traces do not contain inactive cases.}
\begin{tabular}{ c | c| c |c | c | c | c |c | c} 
 \hline
   \textbf{Name} & $\mathcal{T}_0$ & $\mathcal{T}_{30}$ & $\mathcal{T}_{50}$ & $\mathcal{T}_{70}$ & $\mathcal{T}_{90}$ & $\mathcal{T}_{hi}$ & $\mathcal{T}_{mi}$ & $\mathcal{T}_{lo}$ \\
 \hline
  \textbf{Mean} & 100 & 75.3 & 67.2 & 57.2 & 56.3 & 82.5 & 74.1 & 51.2  \\ 
 \hline
  \textbf{Stdev} & 0 & 14.8 & 11.3 & 11.7 & 14.8 & 23.3 & 22.3 & 18.3  \\ 
 \hline
\end{tabular}
\end{center}
\end{table}

Three datasets are used in this paper: MNIST \citep{lecun1998gradient}, EMNIST \citep{emnist} and SYNTHETIC$(\alpha, \beta)$ \citep{li2018federated}. We build a two-layer MLP model and a two-convolution-layer CNN model respectively for MNIST and EMNIST, both models are defined by \citet{fedavg}. For SYNTHETIC$(\alpha, \beta)$, we use an ordinary logistic regression model.  All models use the vanilla SGD as local optimizers, with batch sizes of 10 for MNIST and EMNIST, and 20 for SYNTHETIC. When generating non-IID data, we sort the MNIST and EMNIST data by labels so that each device is assigned data from one label chosen uniformly at random. For SYNTHETIC$(\alpha, \beta)$, we vary the parameters $\alpha, \beta$ from 0 to 1. The larger $\alpha, \beta$ are, the less IID the dataset becomes. We use the staircase learning rate $\eta_\tau = \eta_0/\tau$ as adopted in our convergence analysis. The initial $\eta_0$ is 2e-3 for MNIST, 5e-4 for EMNIST, and 1 for SYNTHETIC$(\alpha, \beta)$. Unless otherwise noted, the number of samples at each device follows the Type-I Pareto distribution with the Pareto index of 0.5.

\subsection{Comparison of Aggregation Schemes} \label{ssec:exp_schemes}

\begin{figure}[h] 
    \centering
    \includegraphics[width=0.45\textwidth]{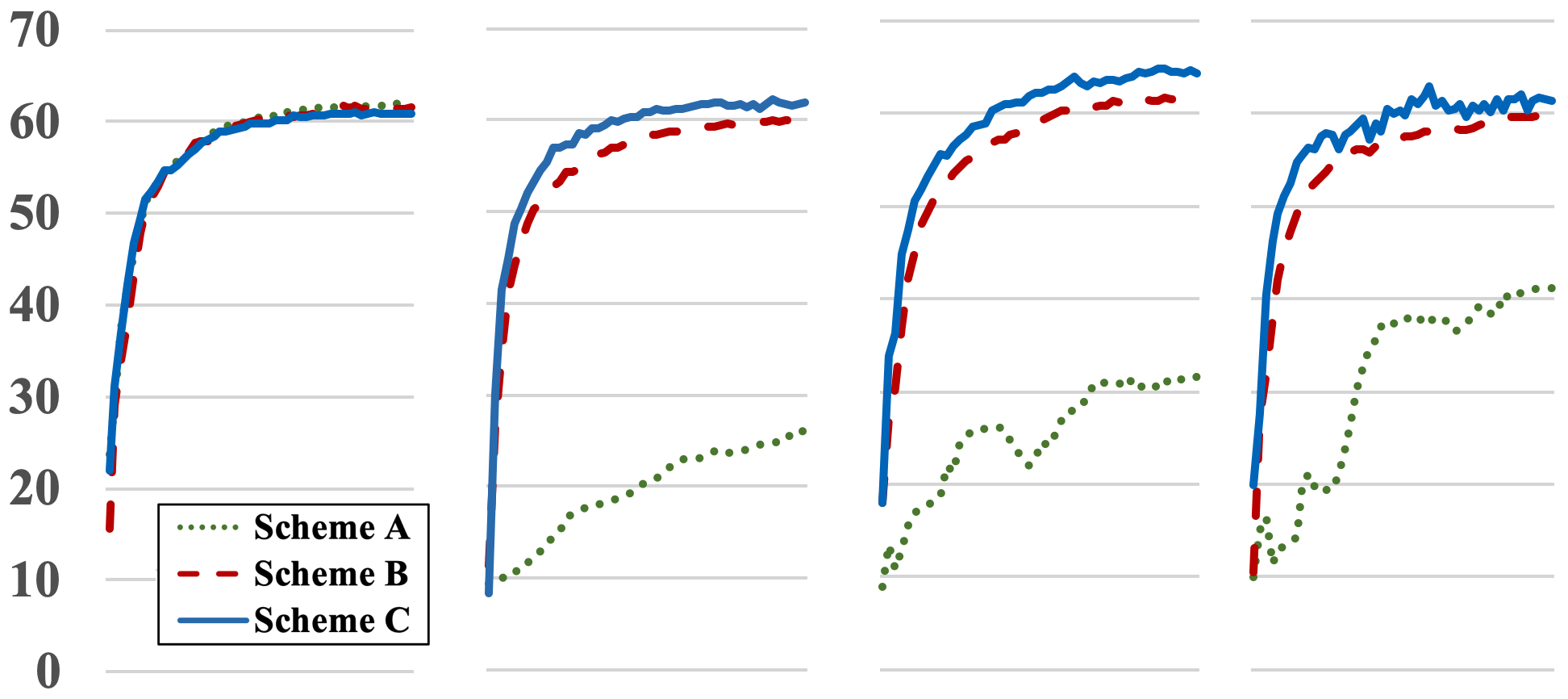}
    \caption{Test accuracy for non-IID EMNIST. Plots from left to right correspond to $|\mathcal{T}|=1,3,5,8$. (increasing device heterogeneity)}\label{fig:emnist_convergence}
\end{figure}

We first examine the effects of the device heterogeneity and the non-IID data distributions on the convergence for each aggregation scheme. We conduct eight sets of experiments where we incrementally increase the number of participation traces to reflect the increasing heterogeneity in device participation. For SYNTHETIC, we use $\alpha=\beta=0$ for the IID case, and $\alpha=\beta=1$ for the non-IID case. We train on 100 devices for MNIST, 62 devices for EMNIST (by merge), and 50 devices for SYNTHETIC$(\alpha, \beta)$. Table \ref{tab:exp_schemes} records the differences in the test accuracies between different aggregation schemes after 200 global epochs. The typical convergence process is depicted in Figure \ref{fig:emnist_convergence}.

As we can see, Scheme C yields the best test accuracy on average. Compared to Schemes A and B, it achieves higher accuracy when devices get more heterogeneous and less IID. This is consistent with our loss bounds in Table \ref{tab:compare}, since Schemes A and B fail to converge to the global optimum in the heterogeneous case with non-IID data. On the other hand, Scheme A performs extremely badly with large $|\mathcal{T}|$. This is because the last few traces contain very few complete rounds, significantly increasing $\mathbb{E}[1/K_\tau]$. Noteworthily, Scheme C is no different from, or even worse than Scheme B in more homogeneous settings, this is consistent with Table \ref{tab:compare} since $\frac{1}{\mathbb{E}[s_\tau]} \leq \mathbb{E}[\frac{1}{s_\tau}]$.  When the traces contain inactive devices ($|\mathcal{T}| \geq 6$), Scheme C becomes less stable due to the variance introduced by $I_t$ in Corollary \ref{cor:compare}.

\begin{table}[h]
\setlength{\tabcolsep}{3pt}
\begin{center}
\caption{The \% improvement in the test accuracies of Scheme B w.r.t. Schemes A(left numbers) and Scheme C w.r.t. Scheme B (right numbers). $|\mathcal{T}|=j$ represents using the first $j$ traces in Table \ref{tab:traces}.} \label{tab:exp_schemes}
\caption*{(a) MNIST Data}
\vspace{-0.5em}
\begin{tabular}{ C{1.0cm} | C{1.6cm}| C{1.6cm} |C{1.6cm} | C{1.6cm} } 
 \hline
   \textbf{$|\mathcal{T}|$} & 1 & 2 & 3 & 4  \\
 \hline
 \textbf{IID} & 
 \makebox[0.7cm][c]{-0.6} |\makebox[0.7cm][c]{0.3} & \makebox[0.7cm][c]{1.9} |\makebox[0.7cm][c]{0.1} & \makebox[0.7cm][c]{5.6} |\makebox[0.7cm][c]{0.1} & \makebox[0.7cm][c]{8.3} |\makebox[0.7cm][c]{0.7} \\ 
 \hline
  \textbf{NIID} & \makebox[0.7cm][c]{0.2} |\makebox[0.7cm][c]{-0.3} & \makebox[0.7cm][c]{9.5} |\makebox[0.7cm][c]{1.8} & \makebox[0.7cm][c]{19.3} |\makebox[0.7cm][c]{1.6} & \makebox[0.7cm][c]{33.8} |\makebox[0.7cm][c]{3.3}  \\ 
 \hline
\end{tabular}

\begin{tabular}{ C{1.0cm} | C{1.6cm}| C{1.6cm} |C{1.6cm} | C{1.6cm}} 
 \hline
   \textbf{$|\mathcal{T}|$} & 5 & 6 & 7 & 8 \\
 \hline
 \textbf{IID} & \makebox[0.7cm][c]{10.4} |\makebox[0.7cm][c]{2.6} & \makebox[0.7cm][c]{14.0} |\makebox[0.7cm][c]{2.2} & \makebox[0.7cm][c]{5.8} |\makebox[0.7cm][c]{1.9} & \makebox[0.7cm][c]{11.8} |\makebox[0.7cm][c]{2.4}\\ 
 \hline
 \textbf{NIID} & \makebox[0.7cm][c]{33.2} |\makebox[0.7cm][c]{3.2} & \makebox[0.7cm][c]{28.7} |\makebox[0.7cm][c]{6.2} & \makebox[0.7cm][c]{36.0} |\makebox[0.7cm][c]{3.6} & \makebox[0.7cm][c]{43.4} |\makebox[0.7cm][c]{6.9}  \\ 
 \hline
\end{tabular}
\end{center}

\setlength{\tabcolsep}{3pt}
\begin{center}
\caption*{(b) EMNIST Data}
\vspace{-0.5em}
\begin{tabular}{ C{1.0cm} | C{1.6cm}| C{1.6cm} |C{1.6cm} | C{1.6cm} } 
 \hline
   \textbf{$|\mathcal{T}|$} & 1 & 2 & 3 & 4  \\
 \hline
 \textbf{IID} & 
 \makebox[0.7cm][c]{0.7} |\makebox[0.7cm][c]{-0.6} & \makebox[0.7cm][c]{0.9} |\makebox[0.7cm][c]{0.1} & \makebox[0.7cm][c]{4.2} |\makebox[0.7cm][c]{0.7} & \makebox[0.7cm][c]{4.8} |\makebox[0.7cm][c]{1.0} \\ 
 \hline
  \textbf{NIID} & \makebox[0.7cm][c]{-0.1} |\makebox[0.7cm][c]{-0.7} & \makebox[0.7cm][c]{17.0} |\makebox[0.7cm][c]{-2.0} & \makebox[0.7cm][c]{34.2} |\makebox[0.7cm][c]{1.8} & \makebox[0.7cm][c]{37.9} |\makebox[0.7cm][c]{4.8}  \\ 
 \hline
\end{tabular}

\begin{tabular}{ C{1.0cm} | C{1.6cm}| C{1.6cm} |C{1.6cm} | C{1.6cm}} 
 \hline
   \textbf{$|\mathcal{T}|$} & 5 & 6 & 7 & 8 \\
 \hline
 \textbf{IID} & \makebox[0.7cm][c]{6.9} |\makebox[0.7cm][c]{1.1} & \makebox[0.7cm][c]{6.6} |\makebox[0.7cm][c]{1.2} & \makebox[0.7cm][c]{4.0} |\makebox[0.7cm][c]{1.5} & \makebox[0.7cm][c]{7.6} |\makebox[0.7cm][c]{1.2}\\ 
 \hline
 \textbf{NIID} & \makebox[0.7cm][c]{30.2} |\makebox[0.7cm][c]{2.5} & \makebox[0.7cm][c]{22.5} |\makebox[0.7cm][c]{3.0} & \makebox[0.7cm][c]{25.3} |\makebox[0.7cm][c]{2.2} & \makebox[0.7cm][c]{18.6} |\makebox[0.7cm][c]{1.8}  \\ 
 \hline
\end{tabular}
\end{center}

\setlength{\tabcolsep}{3pt}
\begin{center}
\caption*{(c) SYNTHETIC Data}
\vspace{-0.5em}
\begin{tabular}{ C{1.0cm} | C{1.6cm}| C{1.6cm} |C{1.6cm} | C{1.6cm} } 
 \hline
   \textbf{$|\mathcal{T}|$} & 1 & 2 & 3 & 4  \\
 \hline
 \textbf{IID} & 
 \makebox[0.7cm][c]{-0.6} |\makebox[0.7cm][c]{0.5} & \makebox[0.7cm][c]{2.1} |\makebox[0.7cm][c]{0.1} & \makebox[0.7cm][c]{6.6} |\makebox[0.7cm][c]{0.0} & \makebox[0.7cm][c]{9.0} |\makebox[0.7cm][c]{0.7} \\ 
 \hline
  \textbf{NIID} & \makebox[0.7cm][c]{0.1} |\makebox[0.7cm][c]{-0.4} & \makebox[0.7cm][c]{9.6} |\makebox[0.7cm][c]{1.5} & \makebox[0.7cm][c]{22.2} |\makebox[0.7cm][c]{1.8} & \makebox[0.7cm][c]{38.2} |\makebox[0.7cm][c]{3.2}  \\ 
 \hline
\end{tabular}

\begin{tabular}{ C{1.0cm} | C{1.6cm}| C{1.6cm} |C{1.6cm} | C{1.6cm}} 
 \hline
   \textbf{$|\mathcal{T}|$} & 5 & 6 & 7 & 8 \\
 \hline
 \textbf{IID} & \makebox[0.7cm][c]{11.6} |\makebox[0.7cm][c]{3.0} & \makebox[0.7cm][c]{16.4} |\makebox[0.7cm][c]{2.5} & \makebox[0.7cm][c]{6.3} |\makebox[0.7cm][c]{1.9} & \makebox[0.7cm][c]{14.4} |\makebox[0.7cm][c]{2.8}\\ 
 \hline
 \textbf{NIID} & \makebox[0.7cm][c]{33.3} |\makebox[0.7cm][c]{3.9} & \makebox[0.7cm][c]{30.5} |\makebox[0.7cm][c]{7.9} & \makebox[0.7cm][c]{37.9} |\makebox[0.7cm][c]{4.5} & \makebox[0.7cm][c]{41.6} |\makebox[0.7cm][c]{8.0}  \\ 
 \hline
\end{tabular}
\end{center}
\end{table}

\subsection{Effectiveness of Fast-Reboot} \label{ssec:exp_arrival}
We now investigate the effectiveness of the fast-reboot method described in Section \ref{ssec:fast_reboot}. The experiments involve $N-1$ existing devices, and the arriving device joins at $\tau_0$. As is discussed in Section \ref{ssec:fast_reboot}, the method makes no difference when data distribution is IID. We thus only consider non-IID cases. We set $N=10$ for MNIST and EMNIST (balanced) and $N=30$ for SYNTHETIC$(1, 1)$. To avoid the interference brought by inactive devices, for this experiment we only use the first five traces in Table \ref{tab:traces}, and we adopt Scheme C as the aggregation method. All devices are given the same number of samples for fair comparison.

\begin{figure}[h] 
    \centering
    \includegraphics[width=0.48\textwidth]{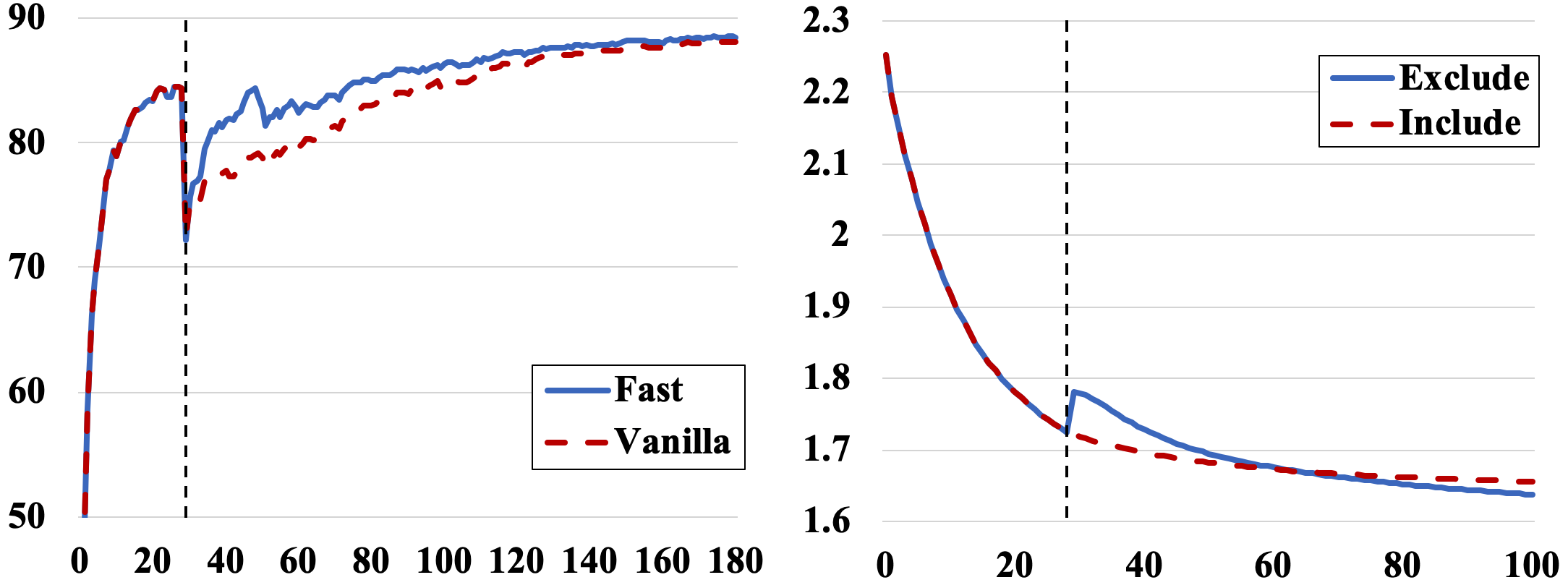}
    \caption{Evolution of the test accuracy (left) and loss (right) under device arrival (left, MNIST) and departure (right, SYNTHETIC) cases. The dashed vertical lines indicate the arriving (departing) time $\tau_0$. After $\tau_0$, except for the ``include'' option, models are tested with new datasets that include (exclude) holdout data from the arriving (departing) device.}\label{fig:exp_arrival_departure}
\end{figure}

\begin{figure}[h] 
    \centering
    \includegraphics[width=0.45\textwidth]{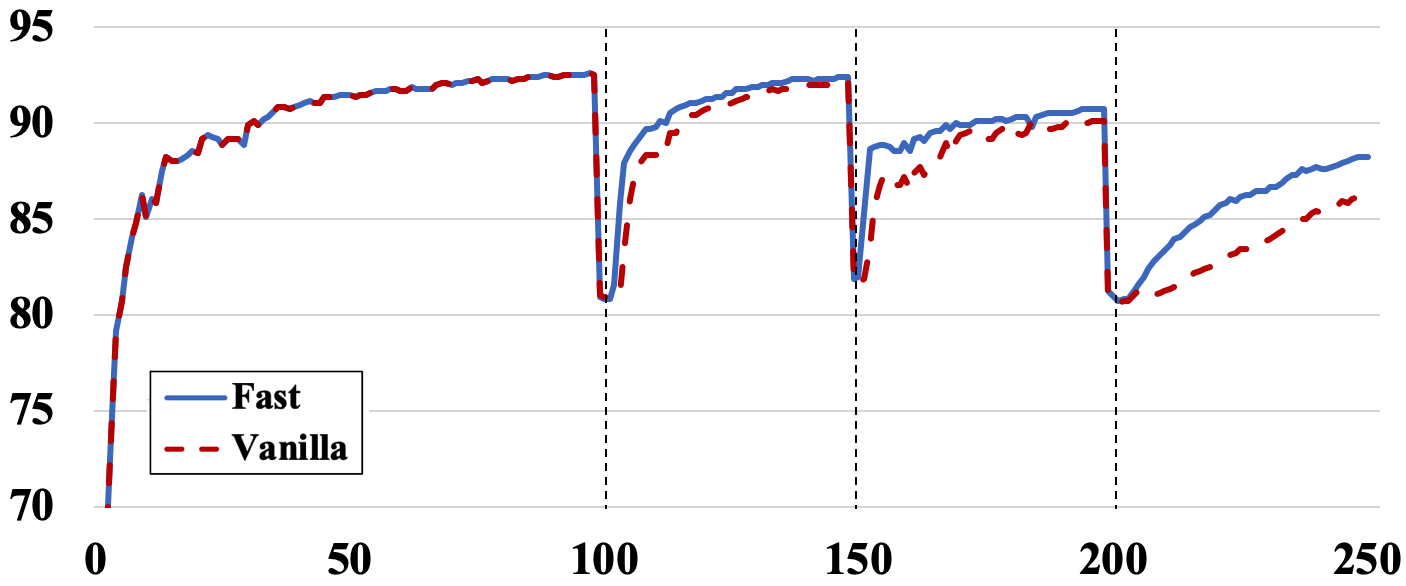}
    \caption{Test accuracy with and without fast-reboot for multiple arrivals for non-IID MNIST. The test dataset is updated every time a new device arrives to include its holdout data. The vertical dashed lines indicate the time the device arrives.}\label{fig:exp_multi_arrival}
\end{figure}

\begin{table}[h]
\setlength{\tabcolsep}{3pt}
\begin{center}
\caption{\label{tab:fast-reboot} The number of global epochs after the arriving time $\tau_0$ until the test accuracy bounces back to that at $\tau_0-1$. Left: fast reboot. Right: vanilla reboot.}
\begin{tabular}{  C{2.5cm}|  C{1.1cm} | C{1.2cm} |  C{1.2cm} |  C{1.2cm} } 
 \hline
  $\boldsymbol{\tau_0}$ & 10 & 30 & 50 & 70  \\
 \hline
  \textbf{MNIST} & \makebox[0.4cm][c]{4} |\makebox[0.4cm][c]{4} & \makebox[0.4cm][c]{22} |\makebox[0.4cm][c]{27} & \makebox[0.4cm][c]{55} |\makebox[0.4cm][c]{63} & \makebox[0.4cm][c]{59} |\makebox[0.4cm][c]{66}     \\ 
 \hline
  \textbf{EMNIST} & \makebox[0.4cm][c]{4} |\makebox[0.4cm][c]{3} & \makebox[0.4cm][c]{11} |\makebox[0.4cm][c]{12} & \makebox[0.4cm][c]{14} |\makebox[0.4cm][c]{19} & \makebox[0.4cm][c]{21} |\makebox[0.4cm][c]{24}   \\ 
 \hline
 \textbf{SYNTHETIC} & \makebox[0.4cm][c]{1} |\makebox[0.4cm][c]{1} & \makebox[0.4cm][c]{4} |\makebox[0.4cm][c]{6} & \makebox[0.4cm][c]{7} |\makebox[0.4cm][c]{12} & \makebox[0.4cm][c]{3} |\makebox[0.4cm][c]{8}    \\ 
 \hline
\end{tabular}
\end{center}
\end{table}

When the device arrives, we increase the learning rate to $\eta_0/(\tau - \tau_0)$. The aggregation coefficient of the arriving device $l$ is boosted to $p_\tau^l = 3p^l$ initially, and decays to $p^l$ by $O(\tau^{-2})$. Table \ref{tab:fast-reboot} records the number of global epochs it takes to recover to the accuracy level before the arrival. Fast-reboot consistently achieves faster rebound, and works better for late arrivals as we expect. EMNIST-CNN enjoys less improvement from fast-reboot because CNN models converge more slowly than MLP and logistic regression models. Thus, at the moment new devices arrive, EMNIST models have not fully converged to the old optima, degrading the effectiveness of fast-reboot as per Corollary \ref{coro:fast_reboot}. The typical fast-reboot process is shown in Figure \ref{fig:exp_arrival_departure}.

Next we study the situation when multiple devices arrive in a row. Figure \ref{fig:exp_multi_arrival} shows the training process for MNIST data. Every time a device arrives, we increase the learning rate as per Corollary \ref{coro:convergence_shift}. Initially, seven devices are in the training. After 100 global epochs, the remaining three devices arrive at 50 epoch intervals, without waiting for the model to fully converge. From Figure \ref{fig:exp_multi_arrival}, the fast-reboot trick accelerates the convergence for every device arrival. 

\subsection{Model Applicability upon Departures} \label{ssec:exp_departure}
The right plot in Figure \ref{fig:exp_arrival_departure} shows the typical change of the test loss after the device departs. We use the same setting as in Section \ref{ssec:exp_arrival}. As is predicted in Section \ref{ssec:applicability}, an objective shift (`exclude') initially increases the test loss. But eventually,  the two curves cross and excluding the device becomes more beneficial.

Table \ref{tab:exp_departure} summarizes the number of global epochs it takes for the curves to cross with SYNTHETIC$(\alpha,\beta)$. As we can see, the values increase with $\tau_0$ and the non-IID metric $(\alpha,\beta)$, confirming Corollary \ref{cor:departure}.

\begin{table}[h]
\setlength{\tabcolsep}{3pt}
\begin{center}
\caption{\label{tab:exp_departure} The number of global epochs after the departing time $\tau_0$ until the test losses coincide for including and excluding options. The rows correspond to three choices of parameters $(\alpha,\beta)$ in SYNTHETIC$(\alpha,\beta)$.}
\begin{tabular}{  C{1.2cm}|  C{0.5cm} | C{0.5cm} |  C{0.5cm} | C{0.5cm} | C{0.5cm} | C{0.5cm} | C{0.5cm} | C{0.5cm} | C{0.5cm} } 
 \hline
  $\boldsymbol{\tau_0}$  & 10 & 15 & 20 & 25 & 30 & 35 & 40 & 45 & 50  \\
 \hline
  $\boldsymbol{(.1,.1)}$  & 2 & 5 & 3 & 3 & 9 & 3 & 10 & 26 & 40\\ 
 \hline
  $\boldsymbol{(.5,.5)}$  & 1 & 3 & 9 & 14 & 13 & 7 & 12 & 36 & 34\\ 
 \hline
 $\boldsymbol{(1.,1.)}$ & 10 & 9 & 27 & 18 & 34 & 17 & 28 & 62 & 77\\ 
 \hline
\end{tabular}
\end{center}
\end{table}

\section{Conclusion and Future Work} \label{sec:conclusion}
This paper extends the federated learning paradigm to incorporate more flexible device participation. The analysis shows that incomplete local device updates can be utilized by scaling the corresponding aggregation coefficients, and a mild degree of device inactivity will not impact the convergence. Further investigation reveals how the convergence relates to heterogeneity in both the data and the device participation. The paper also proposes techniques to fast-reboot the training after new devices arrive, and provides an analytical criterion on when to exclude a departing device. In the future work, we will analyze groups of arrivals or departures, and investigate the possibility for users to dynamically update their datasets during the training.

\subsubsection*{Acknowledgements}
This research was partially supported by the CMU CyLab IoT Initiative, and NSF CNS-1909306, CNS-1751075.

\bibliographystyle{plainnat}
\bibliography{0-main.bib}
\ifcase\TechReport\or
\newpage
\appendix

\newpage
\onecolumn
\section{Proof of Theorems and Corollaries}
\subsection{Proof of Theorem \ref{tm:convergence0}}
\subsubsection{Equivalent View}
For ease of the analysis, we introduce for each client $k$ and each global round $\tau$ a sequence of virtual variables $\alpha_{\tau E}^k, \alpha_{\tau E + 1}^k, \dots, \alpha_{(\tau+1)E - 1}^k$. Here each $\alpha_t^k \in \{0,1\}$ and $\sum_{i=0}^E \alpha_{\tau E + i}^k = s_\tau^k$. Since $s_\tau^k$ is a random variable, $\alpha_t^k$'s are also random variables, and the distributions of $\alpha_t^k$'s determine the distribution of $s_\tau^k$. For example, if $\alpha_t^k \overset{iid}{\sim} \textrm{Bernoulli}(p)$, then $s_\tau^k \sim \textrm{Bin}(E,p)$. In general, we do not make any assumption on the distributions and correlations of $\alpha_t^k$'s. Our results are thus valid for any realization of $s_\tau^k$. 

With the definition of $\alpha_t^k$'s, we can rewrite (\ref{eq:local_sgd_raw})(\ref{eq:aggregation_raw}) as:
\begin{equation} \label{eq:local_sgd}
    w_{\tau E + i + 1}^k = w_{\tau E + i}^k - \eta_{\tau} g_{\tau E +i}^k\alpha_{\tau E + i}^k
\end{equation}
\begin{equation} \label{eq:aggregation}
    w_{(\tau+1)E}^{\mathcal{G}} = w_{\tau E}^{\mathcal{G}} - \sum_{k=1}^N p_\tau^k \sum_{i=0}^{E} \eta_{\tau}g_{\tau E + i}^k\alpha_{\tau E + i}^k
\end{equation}

Note that $w_t^\mathcal{G}$ is visible only when $t$ is a multiple of $E$. To generalize it to arbitrary $t$, we define $\bar{w}_t$ such that $\bar{w}_0 = w_0^\mathcal{G}$, and
\begin{equation}\label{eq:w_bar}
\bar{w}_{\tau E + i + 1} = \bar{w}_{\tau E + i} - \eta_{\tau}\sum_{k=1}^N p_\tau^k g_{\tau E + i}^k \alpha_{\tau E + i}^k
\end{equation}
Note that $\bar{w}_{\tau E + i} = \sum_{k=1}^N p_\tau^k w_{\tau E + i}^k$ only if $\sum_{k=1}^N p_\tau^k = 1$, which generally does not hold.

\begin{lemma}
For any $\tau$, $\bar{w}_{\tau E} = w_{\tau E}^\mathcal{G}$.
\end{lemma}

\begin{proof}
We will prove by induction. By definition, $\bar{w}_0 = w_0^\mathcal{G}$. Suppose $\bar{w}_{\tau E} = w_{\tau E}^\mathcal{G}$, then
\begin{equation}
\begin{split}
    \bar{w}_{(\tau+1) E} &= \bar{w}_{(\tau+1) E - 1} - \eta_{\tau}\sum_{k=1}^N p_\tau^k g_{(\tau+1) E - 1}^k \alpha_{(\tau+1) E - 1}^k \\
    &= \dots = \bar{w}_{\tau E} - \sum_{i=0}^{E-1}\eta_{\tau}\sum_{k=1}^N p_\tau^k g_{\tau E + i}^k \alpha_{\tau E + i}^k\\
    &= w_{\tau E}^\mathcal{G} - \sum_{k=1}^N p_\tau^k \sum_{i=0}^{E-1} \eta_{\tau}g_{\tau E + i}^k\alpha_{\tau E + i}^k = w_{(\tau + 1)E}^\mathcal{G}
\end{split}
\end{equation}
\end{proof}
Thus, in the following analysis we will just use $\bar{w}_t$ to denote the global weight.

\subsubsection{Key Lemmas}
We first present a couple of important lemmas:

\begin{lemma} \label{lm:grad_var}
\begin{equation}
    \mathbb{E}_\xi \| \sum_{k=1}^N p_\tau^k (g_t^k - \bar{g}_t^k) \|^2 \leq \sum_{k=1}^N (p_\tau^k)^2 \sigma_k^2
\end{equation}
\end{lemma}

\begin{proof}
\begin{equation}
    \| \sum_{k=1}^N p_\tau^k (g_t^k - \bar{g}_t^k) \|^2 = \sum_{k=1}^N \| p_\tau^k (g_t^k - \bar{g}_t^k) \|^2 + \sum_{j\neq k}  p_\tau^k p_\tau^j \langle g_t^k - \bar{g}_t^k, g_t^j - \bar{g}_t^j \rangle
\end{equation}
Since each client is running independently, the covariance
\begin{equation}
    \mathbb{E}_\xi \langle g_t^k - \bar{g}_t^k, g_t^j - \bar{g}_t^j \rangle = 0
\end{equation}
Thus,
\begin{equation}
    \mathbb{E}_\xi \| \sum_{k=1}^N p_\tau^k (g_t^k - \bar{g}_t^k) \|^2 = \sum_{k=1}^N \mathbb{E}_\xi \| p_\tau^k (g_t^k - \bar{g}_t^k) \|^2 \leq \sum_{k=1}^N (p_\tau^k)^2 \sigma_k^2
\end{equation}
\end{proof}

\begin{lemma} \label{lm:local_global_div} For $i = 0,\cdots,E-1$ and all $\tau, k$
\begin{equation}
    \mathbb{E}_\xi [\sum_{k=1}^N p_\tau^k \| \bar{w}_{\tau E + i} - w_{\tau E + i}^k \|^2] \leq (E-1)G^2\eta_\tau^2\Big(\sum_{k=1}^Np_\tau^k s_\tau^k + (\sum_{k=1}^N p_\tau^k - 2)_+\sum_{k=1}^N \frac{(p_\tau^k)^2}{p^k}s_\tau^k \Big)
\end{equation}
\end{lemma}

\begin{proof}
Note that $w_{\tau E}^k = \bar{w}_{\tau E}$ for all $k$.
\begin{equation}
\begin{split}
    \| \bar{w}_{\tau E + i} &- w_{\tau E + i}^k \|^2 = \| (\bar{w}_{\tau E + i} - \bar{w}_{\tau E}) - (w_{\tau E + i}^k - \bar{w}_{\tau E}) \|^2 \\
    &= \| \bar{w}_{\tau E + i} - \bar{w}_{\tau E} \|^2 - 2 \langle \bar{w}_{\tau E + i} - \bar{w}_{\tau E}, w_{\tau E + i}^k - \bar{w}_{\tau E} \rangle + \| w_{\tau E + i}^k - \bar{w}_{\tau E} \|^2
\end{split}
\end{equation}
From (\ref{eq:local_sgd})(\ref{eq:w_bar}),

\begin{equation}
\begin{split}
        \sum_{k=1}^N p_\tau^k w_{\tau E + i}^k &= \sum_{k=1}^N p_\tau^k w_{\tau E + i -1}^k - \eta_{\tau}\sum_{k=1}^N p_\tau^k g_{\tau E + i - 1}^k\alpha_{\tau E + i - 1}^k \\
        &= \sum_{k=1}^N p_\tau^k w_{\tau E + i -1}^k + \bar{w}_{\tau E + i} - \bar{w}_{\tau E + i - 1}\\
        &= \dots = \sum_{k=1}^N p_\tau^k w_{\tau E}^k + \bar{w}_{\tau E + i} - \bar{w}_{\tau E}
\end{split}
\end{equation}
Thus,
\begin{equation}
\begin{split}
        &- 2 \sum_{k=1}^N p_\tau^k \langle \bar{w}_{\tau E + i} - \bar{w}_{\tau E}, w_{\tau E + i}^k - \bar{w}_{\tau E} \rangle \\
        =& - 2  \langle \bar{w}_{\tau E + i} - \bar{w}_{\tau E}, \sum_{k=1}^N p_\tau^k w_{\tau E}^k + \bar{w}_{\tau E + i} - \bar{w}_{\tau E} - \sum_{k=1}^N p_\tau^k \bar{w}_{\tau E} \rangle\\
        =& -2 \| \bar{w}_{\tau E + i} - \bar{w}_{\tau E} \|^2
\end{split}
\end{equation}

\begin{equation}\label{eq:lemma2_0}
    \sum_{k=1}^N p_\tau^k \| \bar{w}_{\tau E + i} - w_{\tau E + i}^k \|^2 = (\sum_{k=1}^N p_\tau^k - 2) \| \bar{w}_{\tau E + i} - \bar{w}_{\tau E} \|^2 + \sum_{k=1}^N p_\tau^k \| w_{\tau E + i}^k - \bar{w}_{\tau E} \|^2
\end{equation}

\begin{equation} \label{eq:lemma2_1}
\begin{split}
    &\| \bar{w}_{\tau E + i} - \bar{w}_{\tau E} \|^2 = \| \sum_{j=0}^{i-1} \eta_\tau \sum_{k=1}^N p_\tau^k g_{\tau E + j}^k \alpha_{\tau E + j}^k \|^2 \\
    =& \| \eta_\tau \sum_{k=1}^N p_\tau^k \Big(\sum_{j=0}^{i-1} g_{\tau E + j}^k \alpha_{\tau E + j}^k\Big) \|^2 = \eta_\tau^2 \| \sum_{k=1}^N p^k \Big( \frac{p_\tau^k}{p^k} \sum_{j=0}^{i-1} g_{\tau E + j}^k \alpha_{\tau E + j}^k\Big) \|^2\\
    \leq &\eta_\tau^2 \sum_{k=1}^N \frac{(p_\tau^k)^2}{p^k} \| \sum_{j=0}^{i-1} g_{\tau E + j}^k \alpha_{\tau E + j}^k \|^2
\end{split}
\end{equation}

Here
\begin{equation}
\begin{split}
        \| \sum_{j=0}^{i-1} g_{\tau E + j}^k \alpha_{\tau E + j}^k \|^2 &= \sum_{j=0}^{i-1} \| g_{\tau E + j}^k\alpha_{\tau E + j}^k \|^2 + 2\sum_{p < q} \langle g_{\tau E + p}^k \alpha_{\tau E + p}^k,g_{\tau E + q}^k \alpha_{\tau E + q}^k \rangle \\
        &\leq \sum_{j=0}^{i-1} \| g_{\tau E + j}^k\alpha_{\tau E + j}^k \|^2 + 2\sum_{p< q} \| g_{\tau E + p}^k \alpha_{\tau E + p}^k \| \| g_{\tau E + q}^k \alpha_{\tau E + q}^k \| \\
        &\leq \sum_{j=0}^{i-1} \| g_{\tau E + j}^k\alpha_{\tau E + j}^k \|^2 + \sum_{p < q}\Big(\| g_{\tau E + p}^k \alpha_{\tau E + p}^k \|^2 + \| g_{\tau E + q}^k \alpha_{\tau E + q}^k \|^2 \Big) \\
        &= i\sum_{j=0}^{i-1}\| g_{\tau E + j}^k\alpha_{\tau E + j}^k \|^2
\end{split}
\end{equation}

So
\begin{equation} \label{eq:lemma2_2}
    \mathbb{E}_{\xi} \| \sum_{j=0}^{i-1} g_{\tau E + j}^k \alpha_{\tau E + j}^k \|^2 \leq iG^2 \sum_{j=0}^{i-1} \alpha_{\tau E + j}^k \leq (E-1)G^2s_\tau^k
\end{equation}

Plug (\ref{eq:lemma2_2}) to (\ref{eq:lemma2_1}) we have
\begin{equation} \label{eq:lemma2_3}
\begin{split}
        \mathbb{E}_{\xi} \| \bar{w}_{\tau E + i} - \bar{w}_{\tau E} \|^2 \leq (E-1)G^2\eta_\tau^2\sum_{k=1}^N \frac{(p_\tau^k)^2}{p^k} s_\tau^k
\end{split}
\end{equation}

Similarly
\begin{equation} \label{eq:lemma2_4}
\begin{split}
    \mathbb{E}_\xi\sum_{k=1}^N p_\tau^k \| w_{\tau E + i}^k - \bar{w}_{\tau E} \|^2 = \mathbb{E}_\xi\sum_{k=1}^N p_\tau^k \| \eta_\tau \sum_{j=0}^{i-1} g_{\tau E + j}^k \alpha_{\tau E + j}^k \|^2 \leq (E-1)G^2\eta_\tau^2 \sum_{k=1}^N p_\tau^k s_\tau^k
\end{split}
\end{equation}

Plug (\ref{eq:lemma2_3})(\ref{eq:lemma2_4}) to (\ref{eq:lemma2_0}) we have
\begin{equation}
    \mathbb{E}_\xi [\sum_{k=1}^N p_\tau^k \| \bar{w}_{\tau E + i} - w_{\tau E + i}^k \|^2] \leq (E-1)G^2\eta_\tau^2\Big(\sum_{k=1}^Np_\tau^k s_\tau^k + (\sum_{k=1}^N p_\tau^k - 2)_+\sum_{k=1}^N \frac{(p_\tau^k)^2}{p^k}s_\tau^k \Big)
\end{equation}
\end{proof}

\subsubsection{Bounding $\| \bar{w}_{\tau E + i + 1} - w^* \|^2$}
\begin{equation} \label{eq:|w_bar-w*|}
\begin{split}
    \| \bar{w}_{\tau E + i + 1} - w^* \|^2 &= \|\bar{w}_{\tau E + i} - \eta_\tau \sum_{k=1}^N p_\tau^k \alpha_{\tau E + i}^k g_{\tau E + i}^k - w^* - \eta_{\tau} \sum_{k=1}^N p_\tau^k\alpha_{\tau E + i}^k \bar{g}_{\tau E + i}^k + \eta_{\tau} \sum_{k=1}^N p_\tau^k\alpha_{\tau E + i}^k \bar{g}_{\tau E + i}^k \|^2 \\
    &= \underbrace{\|\bar{w}_{\tau E + i} - w^* - \eta_{\tau} \sum_{k=1}^N p_\tau^k\alpha_{\tau E + i}^k \bar{g}_{\tau E + i}^k\|^2}_{A_1} + \eta_\tau^2 \| \sum_{k=1}^Np_\tau^k\alpha_{\tau E + i}^k(\bar{g}_{\tau E + i}^k - g_{\tau E + i}^k) \|^2 \\
    &+ \underbrace{2\eta_\tau \langle \bar{w}_{\tau E + i} - w^* - \eta_{\tau} \sum_{k=1}^Np_\tau^k\alpha_{\tau E + i}^k\bar{g}_{\tau E + i}^k, \sum_{k=1}^N p_\tau^k\alpha_{\tau E + i}^k (\bar{g}_{\tau E + i}^k - g_{\tau E + i}^k) \rangle}_{A_2}
\end{split}
\end{equation}

Since $\mathbb{E}_\xi [g_{\tau E + i}^k] = \bar{g}_{\tau E + i}^k$, we have $\mathbb{E}_{\xi} [A_2] = 0$. We then bound $A_1$.
\begin{equation} \label{eq:A1_0}
\begin{split}
        &A_1 = \|\bar{w}_{\tau E + i} - w^* - \eta_{\tau} \sum_{k=1}^N p_\tau^k \alpha_{\tau E + i}^k \bar{g}_{\tau E + i}^k\|^2\\
        =& \| \bar{w}_{\tau E + i} - w^* \|^2 \underbrace{- 2\eta_\tau \langle \bar{w}_{\tau E + i} - w^*, \sum_{k=1}^N p_\tau^k\alpha_{\tau E + i}^k \bar{g}_{\tau E + i}^k \rangle}_{B_1} + \underbrace{\eta_\tau^2 \| \sum_{k=1}^N p_\tau^k\alpha_{\tau E + i}^k \bar{g}_{\tau E + i}^k \|^2}_{B_2}
\end{split}
\end{equation}

Since $F_k$ is $L$-smooth,
\begin{equation}
    \| \alpha_{\tau E + i}^k\bar{g}_{\tau E + i}^k \|^2 \leq 2L (F_k(w_{\tau E + i}^k) - F_k^*)\alpha_{\tau E + i}^k
\end{equation}

By the convexity of $l_2$ norm
\begin{equation} \label{eq:B2}
\begin{split}
        B_2 &= \eta_\tau^2 \| \sum_{k=1}^N p_\tau^k\alpha_{\tau E + i}^k \bar{g}_{\tau E + i}^k \|^2 = \eta_\tau^2\| \sum_{k=1}^N p^k (\frac{p_\tau^k}{p^k}\alpha_{\tau E + i}^k \bar{g}_{\tau E + i}^k) \|^2\\
        &\leq \eta_\tau^2 \sum_{k=1}^N \frac{(p_\tau^k)^2}{p^k} \| \alpha_{\tau E + i}^k\bar{g}_{\tau E + i}^k \|^2 \leq 2L\theta\eta_\tau^2 \sum_{k=1}^N p_\tau^k (F_k(w_{\tau E + i}^k) - F_k^*) \alpha_{\tau E + i}^k
\end{split}
\end{equation}

\begin{equation} \label{eq:B1}
\begin{split}
    B_1 &= - 2\eta_\tau \langle \bar{w}_{\tau E + i} - w^*, \sum_{k=1}^N p_\tau^k\alpha_{\tau E + i}^k \bar{g}_{\tau E + i}^k \rangle = -2\eta_\tau \sum_{k=1}^N p_\tau^k \langle \bar{w}_{\tau E + i} - w^*, \alpha_{\tau E + i}^k \bar{g}_{\tau E + i}^k \rangle \\
    &= -2\eta_\tau \sum_{k=1}^N p_\tau^k \langle \bar{w}_{\tau E + i} - w_{\tau E + i}^k, \alpha_{\tau E + i}^k \bar{g}_{\tau E + i}^k \rangle - 2\eta_\tau \sum_{k=1}^N p_k \langle w_{\tau E + i}^k -w^*, \alpha_{\tau E + i}^k \bar{g}_{\tau E + i}^k\rangle
\end{split}
\end{equation}
Here
\begin{equation} \label{eq:B1_1}
\begin{split}
    &-2\langle\bar{w}_{\tau E + i} - w_{\tau E + i}^k, \alpha_{\tau E + i}^k \bar{g}_{\tau E + i}^k \rangle \leq 2 |\langle \bar{w}_{\tau E + i} - w_{\tau E + i}^k, \alpha_{\tau E + i}^k \bar{g}_{\tau E + i}^k \rangle| \\
    \leq& 2 \alpha_{\tau E + i}^k\| \bar{w}_{\tau E + i} - w_{\tau E + i}^k\| \| \bar{g}_{\tau E + i}^k\| \leq \Big(\frac{1}{\eta_\tau} \| \bar{w}_{\tau E + i} - w_{\tau E + i}^k \|^2 + \eta_\tau \| \bar{g}_{\tau E + i}^k \|^2\Big)\alpha_{\tau E + i}^k
\end{split}
\end{equation}
Since $F_k$ is $\mu$-strongly convex
\begin{equation} \label{eq:B1_2}
    \langle w_{\tau E + i}^k - w^*, \alpha_{\tau E + i}^k\bar{g}_{\tau E + i}^k \rangle \geq \big((F_k(w_{\tau E + i}^k) - F_k(w^*)) + \frac{\mu}{2} \| w_{\tau E + i}^k - w^* \|^2\big)\alpha_{\tau E + i}^k
\end{equation}

Plug (\ref{eq:B1_1})(\ref{eq:B1_2}) to (\ref{eq:B1})
\begin{equation} \label{eq:B1_final}
\begin{split}
        B_1 \leq  \sum_{k=1}^N p_\tau^k \alpha_{\tau E + i}^k \Bigg(\| \bar{w}_{\tau E + i} - w_{\tau E + i}^k \|^2 + \eta_\tau^2 \| \bar{g}_{\tau E + i}^k \|^2 - 2\eta_\tau \big((F_k(w_{\tau E + i}^k) - F_k(w^*)) + \frac{\mu}{2} \| w_{\tau E + i}^k - w^* \|^2\big) \Bigg)
\end{split}
\end{equation}

Plug (\ref{eq:B2})(\ref{eq:B1_final}) to (\ref{eq:A1_0})
\begin{equation}
\begin{split}
    A_1 &\leq \|\bar{w}_{\tau E + i} - w^* \|^2 + 2L\theta\eta_\tau^2 \sum_{k=1}^N p_\tau^k\alpha_{\tau E + i}^k (F_k(w_{\tau E + i}^k) - F_k^*) \\
    &+ \sum_{k=1}^N p_\tau^k \alpha_{\tau E + i}^k \Bigg(\| \bar{w}_{\tau E + i} - w_{\tau E + i}^k \|^2 + \underbrace{\eta_\tau^2 \| \bar{g}_{\tau E + i}^k \|^2}_{\leq  2\eta_\tau^2 L(F_k(w_{\tau E + i}^k) - F_k^*)} - 2\eta_\tau \big((F_k(w_{\tau E + i}^k) - F_k(w^*)) + \frac{\mu}{2} \| w_{\tau E + i}^k - w^* \|^2\big) \Bigg) \\
    &\leq \| \bar{w}_{\tau E + i} - w^* \|^2 - \mu\eta_\tau\sum_{k=1}^N p_\tau^k \alpha_{\tau E + i}^k \| w_{\tau E + i}^k - w^* \|^2 + \sum_{k=1}^N p_\tau^k \alpha_{\tau E + i}^k \| \bar{w}_{\tau E + i} - w_{\tau E + i}^k \|^2\\
    &+\underbrace{2(1+\theta)L\eta_\tau^2\sum_{k=1}^N p_\tau^k\alpha_{\tau E + i}^k (F_k(w_{\tau E + i}^k) - F_k^*) - 2\eta_\tau \sum_{k=1}^N p_\tau^k\alpha_{\tau E + i}^k (F_k(w_{\tau E + i}^k) - F_k(w^*))}_{C}
\end{split}
\end{equation}

\begin{equation}
\begin{split}
        \| w_{\tau E + i}^k - w^* \|^2 &= \| w_{\tau E + i}^k - \bar{w}_{\tau E + i} + \bar{w}_{\tau E + i} - w^* \|^2 \\
        &= \| w_{\tau E + i}^k - \bar{w}_{\tau E + i} \|^2 + \| \bar{w}_{\tau E + i} - w^* \|^2 + 2\langle w_{\tau E + i}^k-\bar{w}_{\tau E + i},\bar{w}_{\tau E + i}-w^*\rangle \\
        &\geq \| w_{\tau E + i}^k - \bar{w}_{\tau E + i} \|^2 + \| \bar{w}_{\tau E + i} - w^* \|^2 - 2\| w_{\tau E + i}^k-\bar{w}_{\tau E + i}\| \|\bar{w}_{\tau E + i} - w^* \| \\
        &\geq \| w_{\tau E + i}^k - \bar{w}_{\tau E + i} \|^2 + \| \bar{w}_{\tau E + i} - w^* \|^2 - (2\| w_{\tau E + i}^k-\bar{w}_{\tau E + i}\|^2 + \frac{1}{2} \|\bar{w}_{\tau E + i} - w^* \|^2)\\
        &= \frac{1}{2} \|\bar{w}_{\tau E + i} - w^* \|^2 - \| w_{\tau E + i}^k - \bar{w}_{\tau E + i} \|^2
\end{split}
\end{equation}
Thus,
\begin{equation} \label{eq:A1}
    A_1 \leq (1-\frac{1}{2}\mu\eta_\tau\sum_{k=1}^N p_\tau^k \alpha_{\tau E + i}^k) \| \bar{w}_{\tau E + i} - w^* \|^2 + (1+\mu \eta_\tau)\sum_{k=1}^N p_\tau^k \alpha_{\tau E + i}^k \| \bar{w}_{\tau E + i} - w_{\tau E + i}^k \|^2 + C
\end{equation}

Let $\gamma_\tau = 2\eta_\tau(1-(1+\theta)L\eta_\tau)$. Assume $\eta_\tau \leq \frac{1}{2(1+\theta)L}$, hence $\eta_\tau \leq \gamma_\tau \leq 2\eta_\tau$.
\begin{equation}
\begin{split}
    C &= -2\eta_\tau(1-(1+\theta)L\eta_\tau)\sum_{k=1}^Np_\tau^k\alpha_{\tau E + i}^k(F_k(w_{\tau E + i}^k)-F_k^*) + 2\eta_\tau\sum_{k=1}^Np_\tau^k\alpha_{\tau E + i}^k(F_k(w^*)-F_k^*)\\
    &= -\gamma_\tau \sum_{k=1}^Np_\tau^k\alpha_{\tau E + i}^k (F_k(w_{\tau E + i}^k)- F_k^* + F_k(w^*)-F_k(w^*)) + 2\eta_\tau\sum_{k=1}^Np_\tau^k\alpha_{\tau E + i}^k(F_k(w^*)-F_k^*)\\
    &= -\gamma_\tau \sum_{k=1}^Np_\tau^k\alpha_{\tau E + i}^k (F_k(w_{\tau E + i}^k)-F_k(w^*)) + (2\eta_\tau-\gamma_\tau)\sum_{k=1}^Np_\tau^k\alpha_{\tau E + i}^k(F_k(w^*)-F_k^*)\\
    &\leq \underbrace{-\gamma_\tau \sum_{k=1}^Np_\tau^k\alpha_{\tau E + i}^k (F_k(w_{\tau E + i}^k)-F_k(w^*))}_{D} + 2(1+\theta)L\eta_\tau^2\sum_{k=1}^Np_\tau^k\alpha_{\tau E + i}^k\Gamma_k\\
\end{split}
\end{equation}

Next we bound $D$
\begin{equation}
\begin{split}
    &\sum_{k=1}^Np_\tau^k\alpha_{\tau E + i}^k (F_k(w_{\tau E + i}^k)- F_k(w^*)) = \sum_{k=1}^N p_\tau^k\alpha_{\tau E + i}^k ( F_k(w_{\tau E + i}^k) - F_k(\bar{w}_{\tau E + i})) + \sum_{k=1}^N p_\tau^k\alpha_{\tau E + i}^k (F_k(\bar{w}_{\tau E + i})-F_k(w^*)) \\
    &\geq \sum_{k=1}^N p_\tau^k\alpha_{\tau E + i}^k \langle \nabla F_k(\bar{w}_{\tau E + i}), w_{\tau E + i}^k - \bar{w}_{\tau E + i} \rangle + \sum_{k=1}^N p_\tau^k\alpha_{\tau E + i}^k (F_k(\bar{w}_{\tau E + i})-F_k(w^*)) \\
    &\geq -\sum_{k=1}^N p_\tau^k\alpha_{\tau E + i}^k  \| \nabla F_k(\bar{w}_{\tau E + i}) \| \| w_{\tau E + i}^k - \bar{w}_{\tau E + i}\| + \sum_{k=1}^N p_\tau^k\alpha_{\tau E + i}^k (F_k(\bar{w}_{\tau E + i})-F_k(w^*)) \\
    &\geq - \frac{1}{2}\sum_{k=1}^N p_\tau^k\alpha_{\tau E + i}^k  (\eta_\tau \underbrace{\| \nabla F_k(\bar{w}_{\tau E + i}) \|^2}_{\leq 2L (F_k(\bar{w}_{\tau E + i})-F_k^*)} + \frac{1}{\eta_\tau} \| w_{\tau E + i}^k - \bar{w}_{\tau E + i} \|^2) + \sum_{k=1}^N p_\tau^k\alpha_{\tau E + i}^k (F_k(\bar{w}_{\tau E + i})-F_k(w^*)) \\
    &\geq - \sum_{k=1}^N p_\tau^k\alpha_{\tau E + i}^k \Big(\eta_\tau L   (F_k(\bar{w}_{\tau E + i}) - F_k^*) + \frac{1}{2\eta_\tau}  \| w_{\tau E + i}^k - \bar{w}_{\tau E + i} \|^2 \Big) + \sum_{k=1}^N p_\tau^k\alpha_{\tau E + i}^k (F_k(\bar{w}_{\tau E + i})-F_k(w^*))
\end{split}
\end{equation}
Thus,
\begin{equation} \label{eq:C}
\begin{split}
        C &\leq \gamma_\tau\sum_{k=1}^N p_\tau^k\alpha_{\tau E + i}^k (\eta_\tau L  \underbrace{(F_k(\bar{w}_{\tau E + i})-F_k^*)}_{F_k(\bar{w}_{\tau E + i})-F_k(w^*) + F_k(w^*) - F_k^*}+\frac{1}{2\eta_\tau}\| w_{\tau E + i}^k - \bar{w}_{\tau E + i} \|^2) \\
        &\;\;\;\;\;\;\;- \gamma_\tau \sum_{k=1}^N p_\tau^k\alpha_{\tau E + i}^k (F_k(\bar{w}_{\tau E + i})-F_k(w^*)) + 2(1+\theta)L\eta_\tau^2 \sum_{k=1}^N p_\tau^k \alpha_{\tau E + i} \Gamma_k \\
        &= \gamma_\tau (\eta_\tau L -1) \sum_{k=1}^N p_\tau^k\alpha_{\tau E + i}^k (F_k(\bar{w}_{\tau E + i}) - F_k(w^*)) + \underbrace{\frac{\gamma_\tau}{2\eta_\tau}}_{\leq 1}\sum_{k=1}^N p_\tau^k\alpha_{\tau E + i}^k \| w_{\tau E + i}^k - \bar{w}_{\tau E + i} \|^2\\
        &\;\;\;\;\;\;\;+ 2(1+\theta)L\eta_\tau^2 \sum_{k=1}^N p_\tau^k \alpha_{\tau E + i} \Gamma_k + \underbrace{\gamma_\tau}_{\leq 2\eta_\tau}\eta_\tau L \sum_{k=1}^N p_\tau^k\alpha_{\tau E + i}^k \Gamma_k\\
        &\leq \gamma_\tau (\eta_\tau L -1) \sum_{k=1}^N p_\tau^k\alpha_{\tau E + i}^k (F_k(\bar{w}_{\tau E + i}) - F_k(w^*)) + \sum_{k=1}^N p_\tau^k\alpha_{\tau E + i}^k \| w_{\tau E + i}^k - \bar{w}_{\tau E + i} \|^2\\
        &\;\;\;\;\;\;\; + 2(2+\theta)L\eta_\tau^2 \sum_{k=1}^N p_\tau^k\alpha_{\tau E + i}^k \Gamma_k
\end{split}
\end{equation}

Plug (\ref{eq:C}) to (\ref{eq:A1}) we have
\begin{equation} \label{eq:A1_final}
\begin{split}
        A_1 &\leq \| \bar{w}_{\tau E + i} - w^* \|^2 - \mu\eta_\tau\sum_{k=1}^N p_\tau^k \alpha_{\tau E + i}^k \| w_{\tau E + i}^k - w^* \|^2 + 2\sum_{k=1}^N p_\tau^k\alpha_{\tau E + i}^k \| \bar{w}_{\tau E + i} - w_{\tau E + i}^k \|^2 \\
        & + 2(2+\theta)L\eta_\tau^2 \sum_{k=1}^N p_\tau^k\alpha_{\tau E + i}^k \Gamma_k + \gamma_\tau (\eta_\tau L -1) \sum_{k=1}^N p_\tau^k\alpha_{\tau E + i}^k (F_k(\bar{w}_{\tau E + i}) - F_k(w^*))
\end{split}
\end{equation}

Plug (\ref{eq:A1_final}) to (\ref{eq:|w_bar-w*|}), 
\begin{equation} \label{eq:recursion0}
\begin{split}
    &\| \bar{w}_{\tau E + i + 1} - w^* \|^2 \leq (1-\frac{1}{2}\mu\eta_\tau\sum_{k=1}^N p_\tau^k \alpha_{\tau E + i}^k) \| \bar{w}_{\tau E + i} - w^* \|^2 \\
    +&\eta_\tau^2 \| \sum_{k=1}^Np_\tau^k\alpha_{\tau E + i}^k(\bar{g}_{\tau E + i}^k - g_{\tau E + i}^k) \|^2 + \underbrace{(2+\mu\eta_\tau)}_{\leq 2 + \frac{\mu}{2(1+\theta)L}}\sum_{k=1}^N p_\tau^k\alpha_{\tau E + i}^k \| \bar{w}_{\tau E + i} - w_{\tau E + i}^k \|^2\\
    +& 2(2+\theta)L\eta_\tau^2 \sum_{k=1}^N p_\tau^k\alpha_{\tau E + i}^k \Gamma_k + \underbrace{\gamma_\tau (1 - \eta_\tau L )}_{\leq 2\eta_\tau} \sum_{k=1}^N p_\tau^k\alpha_{\tau E + i}^k (F_k(w^*) - F_k(\bar{w}_{\tau E + i}))
\end{split}
\end{equation}

Define 
\begin{equation}
\begin{split}
        C_{\tau E + i} = (2 + \frac{\mu}{2(1+\theta)L}) \sum_{k=1}^N p_\tau^k\alpha_{\tau E + i}^k \| \bar{w}_{\tau E + i} - w_{\tau E + i}^k \|^2 &+ \| \sum_{k=1}^Np_\tau^k\alpha_{\tau E + i}^k(\bar{g}_{\tau E + i}^k - g_{\tau E + i}^k) \|^2 \\
        &+ 2(2+\theta)L \sum_{k=1}^N p_\tau^k\alpha_{\tau E + i}^k \Gamma_k
\end{split}
\end{equation}

Thus,
\begin{equation} \label{eq:recursion0}
\begin{split}
    &\| \bar{w}_{\tau E + i + 1} - w^* \|^2 \leq (1-\frac{1}{2}\mu\eta_\tau\sum_{k=1}^N p_\tau^k \alpha_{\tau E + i}^k) \| \bar{w}_{\tau E + i} - w^* \|^2 + \eta_\tau^2B_{\tau E + i}\\
    +& 2\eta_\tau \sum_{k=1}^N p_\tau^k\alpha_{\tau E + i}^k ( F_k(w^*) - F_k(\bar{w}_{\tau E + i}))
\end{split}
\end{equation}

Apply the lemmas we have

\begin{equation}
\begin{split}
        \mathbb{E}_\xi &[C_{\tau E + i}] \leq \sum_{k=1}^N (p_\tau^k)^2\alpha_{\tau E + i}^k \sigma_k^2  + 2(2+\theta)L \sum_{k=1}^N p_\tau^k\alpha_{\tau E + i}^k \Gamma_k\\
        &+(2 + \frac{\mu}{2(1+\theta)L})(E-1)G^2\Big(\sum_{k=1}^Np_\tau^k s_\tau^k + (\sum_{k=1}^N p_\tau^k - 2)_+\sum_{k=1}^N \frac{(p_\tau^k)^2}{p^k}s_\tau^k \Big)
\end{split}
\end{equation}

For convenience we write $\Delta_{\tau E + i} = \| \bar{w}_{\tau E + i} - w^* \|^2$, and $\bar{\Delta}_{\tau E + i} = \mathbb{E}[\Delta_{\tau E + i}]$, where the expectation is taken over all random variables up to $\tau E + i$.

\subsubsection{Bounding $\| \bar{w}_{\tau E} - w^* \|$}
Summing from $\tau E$ to $(\tau+1)E$ we have

\begin{equation}
    \sum_{i=1}^{E}\Delta_{\tau E + i}\leq \sum_{i=0}^{E-1} (1-\frac{1}{2}\mu\eta_\tau\sum_{k=1}^N p_\tau^k \alpha_{\tau E + i}^k) \Delta_{\tau E + i} + \eta_\tau^2C_\tau + 2\eta_\tau \sum_{k=1}^N p_\tau^k s_\tau^k ( F_k(w^*) - F_k(\bar{w}_{\tau E + l}))
\end{equation}

where $C_\tau = \sum_{i=0}^{E-1} C_{\tau E + i}$, and $\bar{w}_{\tau E + l} = \text{argmin}_{\bar{w}_{\tau E + i}} \sum_{k=1}^N p_\tau^k \alpha_{\tau E + i}^k F_k(\bar{w}_{\tau E + i})$.

Reorganize it we can get
\begin{equation} \label{eq:recursive_tauE}
\begin{split}
        \Delta_{(\tau + 1) E} &\leq \Delta_{\tau E} - \frac{1}{2}\mu \eta_\tau \sum_{i=0}^{E-1}\sum_{k=1}^N p_\tau^k \alpha_{\tau E + i}^k \Delta_{\tau E + i} + \eta_\tau^2C_\tau + 2\eta_\tau \sum_{k=1}^N p_\tau^k s_\tau^k ( F_k(w^*) - F_k(\bar{w}_{\tau E + l}))
\end{split}
\end{equation}

We then seek to find a lower bound for $\Delta_{\tau E + i}$.
\begin{equation}
\begin{split}
        \sqrt{\Delta_{\tau E + i +1}} &= \| \bar{w}_{\tau E + i + 1} - w^* \| = \| \bar{w}_{\tau E + i + 1} - \bar{w}_{\tau E + i} + \bar{w}_{\tau E + i} - w^* \| \\
        &\leq \| \bar{w}_{\tau E + i + 1} - \bar{w}_{\tau E + i} \| + \sqrt{\Delta_{\tau E + i}}\\
        &= \| \eta_\tau \sum_{k=1}^N p_\tau^k\alpha_{\tau E + i}^k g_{\tau E + i}^k \| + \sqrt{\Delta_{\tau E + i}}
\end{split}
\end{equation}
Define $h_{\tau E + i} = \| \sum_{k=1}^N p_\tau^k\alpha_{\tau E + i}^k g_{\tau E + i}^k \|$.

Thus,
\begin{equation}
\begin{split}
        \sqrt{\Delta_{(\tau+1) E}} &\leq \sqrt{\Delta_{(\tau + 1)E - 1}} + \eta_\tau h_{(\tau + 1)E - 1} \\
        &\leq \dots \leq \sqrt{\Delta_{\tau E + i}} + \sum_{j=i}^{E-1}\eta_\tau h_{\tau E + j}
\end{split}
\end{equation}

\begin{equation}
\begin{split}
        \Delta_{(\tau + 1)E} &\leq \Delta_{\tau E + i} + 2\sqrt{\Delta_{\tau E + i}}(\sum_{j=i}^{E-1}\eta_\tau h_{\tau E + j}) + (\sum_{j=i}^{E-1}\eta_\tau h_{\tau E + j})^2\\
        &\leq 2\Delta_{\tau E + i} + 2(\sum_{j=i}^{E-1}\eta_\tau h_{\tau E + j})^2
\end{split}
\end{equation}

\begin{equation} \label{eq:Delta_lower_bound}
    \Delta_{\tau E + i} \geq \frac{1}{2}\Delta_{(\tau + 1)E} - (\sum_{j=i}^{E-1}\eta_\tau h_{\tau E + j})^2 \geq \frac{1}{2}\Delta_{(\tau + 1)E} - (\sum_{j=0}^{E-1}\eta_\tau h_{\tau E + j})^2
\end{equation}

Plug (\ref{eq:Delta_lower_bound}) to (\ref{eq:recursive_tauE}) we can get
\begin{equation}
\begin{split}
        (1+\frac{1}{4}\mu \eta_\tau \sum_{k=1}^N p_\tau^k s_\tau^k)\Delta_{(\tau + 1) E} \leq \Delta_{\tau E} &+ \frac{1}{2}\mu \eta_\tau^3 \sum_{k=1}^N p_\tau^k s_\tau^k (\sum_{i=0}^{E-1} h_{\tau E + i})^2 + \eta_\tau^2C_\tau \\
        &+ 2\eta_\tau \sum_{k=1}^N p_\tau^k s_\tau^k ( F_k(w^*) - F_k(\bar{w}_{\tau E + l}))
\end{split}
\end{equation}

Define $H_\tau = (\sum_{i=0}^{E-1} h_{\tau E + i})^2$. Apply Lemma \ref{lm:grad_var}, Lemma \ref{lm:local_global_div} and Assumption \ref{asm:G2}, we have

\begin{equation}
\begin{split}
        &\mathbb{E}_\xi[h_{\tau E + i}^2] = \mathbb{E}_\xi\| \sum_{k=1}^N p_\tau^k\alpha_{\tau E + i}^k g_{\tau E + i}^k \|^2 \\
        \leq& \sum_{k=1}^N \frac{(p_\tau^k)^2}{p^k}\mathbb{E}_\xi \|\alpha_{\tau E + i}^kg_{\tau E + i}^k\|^2 \leq \sum_{k=1}^N \frac{(p_\tau^k)^2}{p^k}G^2 \alpha_{\tau E + i}^k
\end{split}
\end{equation}

\begin{equation}
\begin{split}
    &\mathbb{E}_\xi[H_\tau] = \mathbb{E}_\xi [(\sum_{i=0}^{E-1} h_{\tau E + i})^2] \leq \mathbb{E}_\xi [E \sum_{i=0}^{E-1} h_{\tau E + i}^2] \leq EG^2\sum_{k=1}^N\frac{(p_\tau^k)^2}{p^k}s_\tau^k
\end{split}
\end{equation}

\begin{equation}
\begin{split}
        \mathbb{E}_\xi &[C_\tau] = \sum_{i=0}^{E-1}\mathbb{E}_\xi[C_{\tau E + i}] = \sum_{k=1}^N (p_\tau^k)^2s_\tau^k \sigma_k^2  + 2(2+\theta)L \sum_{k=1}^N p_\tau^ks_\tau^k \Gamma_k \\
        &+(2 + \frac{\mu}{2(1+\theta)L})E(E-1)G^2 \Big(\sum_{k=1}^Np_\tau^k s_\tau^k + \theta(\sum_{k=1}^N p_\tau^k - 2)_+\sum_{k=1}^N p_\tau^ks_\tau^k \Big)
\end{split}
\end{equation}

Write $\bar{\Delta}_{\tau E + i} = \mathbb{E}_\xi[\Delta_{\tau E + i}]$,$ \bar{C}_\tau = \mathbb{E}_\xi[C_\tau]$, $\bar{H}_\tau = \mathbb{E}_\xi [(\sum_{i=0}^{E-1} h_{\tau E + i})^2]$, then

\begin{equation} \label{eq:recursive_tauE_2}
\begin{split}
        (1+\frac{1}{4}\mu \eta_\tau \sum_{k=1}^N p_\tau^k s_\tau^k)\bar{\Delta}_{(\tau + 1) E} \leq \bar{\Delta}_{\tau E} &+ \frac{1}{2}\mu \eta_\tau^3 \sum_{k=1}^Np_\tau^k s_\tau^k\bar{H}_\tau + \eta_\tau^2\bar{C}_\tau \\
        &+ 2\eta_\tau \mathbb{E}_\xi \sum_{k=1}^N p_\tau^k s_\tau^k ( F_k(w^*) - F_k(\bar{w}_{\tau E + l}))
\end{split}
\end{equation}

Let $z_\tau = 0$ indicate the event that for all $k$, $\mathbb{E}[p_\tau^ks_\tau^k] = c_\tau p^k$ for come constant $c_\tau$ that does not depend on $k$, otherwise $z_\tau^k = 1$. Note that if $z_\tau = 0$, then $\sum_{k=1}^N p_\tau^k s_\tau^k ( F_k(w^*) - F_k(\bar{w}_{\tau E + l})) = c_\tau(F(w^*)-F(\bar{w}_{\tau E + l})) \leq 0$. Otherwise, we have

\begin{equation}
\begin{split}
    \sum_{k=1}^N p_\tau^k s_\tau^k (F_k(w^*) - F_k(\bar{w}_{\tau E + l})) =& \sum_{k=1}^N p_\tau^k s_\tau^k ( \underbrace{F_k(w^*) - F_k^*}_{\Gamma_k} + \underbrace{F_k^* - F_k(\bar{w}_{\tau E + l}}_{\leq 0}))\\
    \leq& \sum_{k=1}^N p_\tau^k s_\tau^k \Gamma_k
\end{split}
\end{equation}

Put it together
\begin{equation}
    \sum_{k=1}^N p_\tau^k s_\tau^k ( F_k(w^*) - F_k(\bar{w}_{\tau E + l})) \leq z_\tau \sum_{k=1}^N p_\tau^k s_\tau^k \Gamma_k
\end{equation}

Assume $\eta_\tau \leq \frac{4}{\mu E\theta} \leq \frac{4}{\mu \sum_{k=1}^N p_\tau^k s_\tau^k}$, divide both sides with $1+\frac{1}{4}\mu\eta_\tau\sum_{k=1}^N p_\tau^k s_\tau^k$ in (\ref{eq:recursive_tauE_2}) we can get

\begin{equation}
\begin{split}
        \bar{\Delta}_{(\tau + 1)E} &\leq \Big(1- \frac{\frac{1}{4}\mu\eta_\tau\sum_{k=1}^N p_\tau^k s_\tau^k}{1+\frac{1}{4}\mu\eta_\tau\sum_{k=1}^N p_\tau^k s_\tau^k} \Big)\bar{\Delta}_{\tau E} + 2 \eta_\tau^2 \bar{H}_\tau + \eta_\tau^2\bar{C}_\tau \\
        &+ 2 \eta_\tau z_\tau \sum_{k=1}^N p_\tau^k s_\tau^k \Gamma_k\\
        &\leq \Big( 1 - \frac{1}{8}\mu\eta_\tau\sum_{k=1}^N p_\tau^k s_\tau^k \Big) \bar{\Delta}_{\tau E} + \eta_\tau^2 B_\tau + 2 \eta_\tau z_\tau \sum_{k=1}^N p_\tau^k s_\tau^k \Gamma_k
\end{split}
\end{equation}

Note that $p_\tau^k$, $s_\tau^k$ are independent with $\bar{\Delta}_{\tau E}$. Taking expectation over $p_\tau^k$ and $s_\tau^k$ we get

\begin{equation}
\begin{split}
        \mathbb{E}[\bar{\Delta}_{(\tau + 1)E}] \leq \Big( 1 - \frac{1}{8}\mu\eta_\tau\mathbb{E}[\sum_{k=1}^N p_\tau^k s_\tau^k] \Big) \bar{\Delta}_{\tau E} + \eta_\tau^2 \mathbb{E}[B_\tau] + 2 \eta_\tau z_\tau \sum_{k=1}^N \mathbb{E}[p_\tau^k s_\tau^k] \Gamma_k
\end{split}
\end{equation}

\subsection{Proof of Theorem \ref{tm:convergence0}}
When the distributions of $s_\tau^k$ do not change with time, we have $B_\tau = B$. We prove the convergence by induction. Let $\eta_\tau = \frac{8}{\mu \mathbb{E}[\sum_{k=1}^Np_\tau^ks_\tau^k]}\frac{2E}{\tau E + \gamma}$. Initially, $\frac{V_0}{\gamma^2} \geq \mathbb{E}[\bar{\Delta}_0]$. Suppose $\mathbb{E}[\bar{\Delta}_{\tau E}] \leq \frac{M_\tau D + V}{\tau E + \gamma}$, then

\begin{equation}
\begin{split}
        &\mathbb{E}[\bar{\Delta}_{(\tau + 1)E}] \leq \frac{\tau E + \gamma - 2E}{\tau E + \gamma}\frac{M_\tau D + V}{\tau E +\gamma} + \left(\frac{16E}{\mu\mathbb{E}[\sum_{k=1}^Np_\tau^ks_\tau^k]}\right)^2\frac{B}{(\tau E + \gamma)^2} + \frac{\frac{1}{2}z_\tau D}{\tau E + \gamma} \\
        \leq& \frac{\tau E + \gamma - E}{(\tau E + \gamma)^2}\left(M_\tau D + V\right) + \frac{\frac{1}{2}z_\tau D}{\tau E + \gamma} + \underbrace{\left(\frac{16E}{\mu\mathbb{E}[\sum_{k=1}^Np_\tau^ks_\tau^k]}\right)^2\frac{B}{(\tau E + \gamma)^2} - \frac{E(M_\tau D + V)}{(\tau E + \gamma)^2}}_{\leq 0} \\
        \leq& \frac{M_\tau D + V}{(\tau + 1) E + \gamma} + \frac{\frac{1}{2}\frac{\tau E + \gamma + E}{\tau E + \gamma}z_\tau D}{(\tau+1) E + \gamma} \leq \frac{M_{\tau + 1}D + V}{(\tau + 1)E + \gamma}
\end{split}
\end{equation}

Thus $\bar{\Delta}_{(\tau + 1)E } \leq \frac{M_{\tau + 1} D + V}{(\tau + 1)E + \gamma}$.

We can check it satisfies the previous assumptions regarding $\eta_\tau$:

\begin{equation}
\begin{split}
        &\eta_\tau \leq \eta_0 = \frac{16E/(\mu \mathbb{E}[\sum_{k=1}^Np_\tau^ks_\tau^k])}{E + \gamma}\\ 
        \leq& \frac{16E/(\mu \mathbb{E}[\sum_{k=1}^Np_\tau^ks_\tau^k])}{32E(1+\theta)L/(\mu \mathbb{E}[\sum_{k=1}^Np_\tau^ks_\tau^k])} = \frac{1}{2(1+\theta)L}
\end{split}
\end{equation}

\begin{equation}
    \eta_\tau \leq \eta_0 = \frac{16E/(\mu \mathbb{E}[\sum_{k=1}^Np_\tau^ks_\tau^k])}{E + \gamma} \leq \frac{16E/(\mu \mathbb{E}[\sum_{k=1}^Np_\tau^ks_\tau^k])}{4E^2\theta/( \mathbb{E}[\sum_{k=1}^Np_\tau^ks_\tau^k])} = \frac{4}{\mu E\theta}
\end{equation}

\subsubsection{Extension to Time-Varying Distributions}
When the distribution of $s_\tau^k$ vary with time, we can still establish a convergence with slightly different definitions. 

Redefine $\gamma = \max \left\{\frac{32E(1+\theta)L}{\mu \min_\tau \mathbb{E}[\sum_{k=1}^Np_\tau^ks_\tau^k]}, \frac{4E^2\theta}{\min_\tau \mathbb{E}[\sum_{k=1}^Np_\tau^ks_\tau^k]} \right\}, V_\tau = \max\left\{\gamma^2 \mathbb{E}\|w_{0}^{\mathcal{G}} - w^* \|^2, \left(\frac{16E}{\mu}\right)^2\sum_{t=0}^{\tau - 1} \frac{\mathbb{E}[B_t]}{\left(\mathbb{E}[\sum_{k=1}^Np_t^ks_t^k]\right)^2} \right\}$

We now prove by induction that with this definition, we can obtain

\begin{equation}
    \mathbb{E}[\bar{\Delta}_{\tau E}] \leq \frac{M_\tau D}{\tau E + \gamma} + \frac{V_\tau}{\left(\tau E + \gamma\right)^2}
\end{equation}

Let $\eta_\tau = \frac{8}{\mu \mathbb{E}[\sum_{k=1}^Np_\tau^ks_\tau^k]}\frac{2E}{(\tau+1) E + \gamma}$. Initially, $\frac{V_0}{\gamma^2} \geq \mathbb{E}[\bar{\Delta}_0]$. Suppose $\mathbb{E}[\bar{\Delta}_{\tau E}] \leq \frac{M_\tau D}{\tau E + \gamma} + \frac{V_\tau}{\left(\tau E + \gamma\right)^2}$, then

\begin{equation}
\begin{split}
        &\mathbb{E}[\bar{\Delta}_{(\tau + 1)E}] \leq \frac{\tau E + \gamma - E}{(\tau+1) E + \gamma}\left(\frac{M_\tau D}{\tau E + \gamma} + \frac{V_\tau}{\left(\tau E + \gamma\right)^2}\right) + \frac{(16E)^2\mathbb{E}[\bar{B}_\tau+2\bar{H}_\tau]}{(\mu\mathbb{E}[\sum_{k=1}^Np_\tau^ks_\tau^k])^2\left((\tau + 1) E + \gamma\right)^2}+\frac{z_\tau D}{(\tau+1) E + \gamma} \\
        \leq& \frac{(\tau E + \gamma - E)M_\tau D}{(\tau E + \gamma)^2-E^2} + \frac{\tau E + \gamma - E}{(\tau E + \gamma)^2-E^2}\frac{V_\tau}{(\tau+1) E + \gamma} + \frac{(16E)^2\mathbb{E}[\bar{B}_\tau+2\bar{H}_\tau]}{(\mu\mathbb{E}[\sum_{k=1}^Np_\tau^ks_\tau^k])^2\left((\tau + 1) E + \gamma\right)^2} + \frac{z_\tau D}{(\tau + 1) E + \gamma}\\
        \leq& \frac{M_\tau D}{(\tau + 1)E + \gamma} + \frac{V_\tau}{\left((\tau + 1)E + \gamma\right)^2} + \frac{(16E)^2\mathbb{E}[\bar{B}_\tau+2\bar{H}_\tau]}{(\mu\mathbb{E}[\sum_{k=1}^Np_\tau^ks_\tau^k])^2\left((\tau + 1) E + \gamma\right)^2} + \frac{z_\tau D}{(\tau + 1)E + \gamma} \\
        =& \frac{M_{\tau + 1} D}{(\tau + 1)E + \gamma} + \frac{V_{\tau + 1}}{\left((\tau + 1)E + \gamma\right)^2}
\end{split}
\end{equation}

Thus $\bar{\Delta}_{(\tau + 1)E } \leq \frac{M_{\tau + 1} D}{(\tau + 1)E + \gamma} + \frac{V_{\tau + 1}}{\left((\tau + 1)E + \gamma\right)^2}$.

Easy to check previous assumptions regarding $\eta_\tau$ are all satisfied.

\subsection{Proof of Theorem \ref{tm:shift}}
\begin{itemize} [leftmargin=*]
    \item Departure Case: $\Tilde{n} = n - n_l$
    \begin{equation*}
    \begin{split}
        &\| \Tilde{w}^* - w^* \| \leq \frac{2}{\mu} \| \nabla F(\Tilde{w}^*) \| = \frac{2}{\mu} \left\| \nabla F(\Tilde{w}^*) - \underbrace{\nabla \Tilde{F}(\Tilde{w}^*)}_{=0}  \right\| \\
        =& \frac{2}{\mu} \left\| \sum_{k \neq l} (p^k - \Tilde{p}^k) \nabla F_k(\Tilde{w}^*) + p^l\nabla F_l(\Tilde{w}^*) \right\| \\
        =& \frac{2}{\mu} \left\| \sum_{k \neq l} \left(\frac{n_k}{n} - \frac{n_k}{n-n_l} \right) \nabla F_k(\Tilde{w}^*) + p^l\nabla F_l(\Tilde{w}^*) \right\| \\
        =& \frac{2}{\mu} \left\| -\sum_{k\neq l} \left(\frac{n_ln_k}{n(n-n_l)} \right) \nabla F_k(\Tilde{w}^*) + p^l\nabla F_l(\Tilde{w}^*) \right\| \\
        =& \frac{2}{\mu} \left\| -p^l \underbrace{\sum_{k\neq l} \Tilde{p}^k \nabla F_k(\Tilde{w}^*)}_{=\nabla \Tilde{F}(\Tilde{w}^*) = 0} + p^l\nabla F_l(\Tilde{w}^*) \right\| \\
        =& \frac{2p^l}{\mu} \left\| \nabla F_l(\Tilde{w}^*) \right\|  \leq \frac{2p^l}{\mu} \sqrt{2L \left(F_l(\Tilde{w}^*) - F_l^*\right)} = \frac{2\sqrt{2L}}{\mu} p^l\sqrt{\Tilde{\Gamma}_l}
    \end{split}
    \end{equation*}
    
    \item Arrival Case: $\Tilde{n} = n + n_l$
    \begin{equation*}
    \begin{split}
        &\| \Tilde{w}^* - w^* \| = \|w^* -  \Tilde{w}^* \| \leq \frac{2}{\mu} \| \nabla \Tilde{F}(w^*) \| = \frac{2}{\mu} \left\| \nabla \Tilde{F}(w^*) - \underbrace{\nabla F(w^*)}_{=0}  \right\| \\
        =& \dots = \frac{2}{\mu} \left\| -\Tilde{p}^l \underbrace{\sum_{k\neq l} p^k \nabla F_k(w^*)}_{=\nabla F(w^*) = 0} + \Tilde{p}^l\nabla F_l(w^*) \right\| \\
        =& \frac{2\Tilde{p}^l}{\mu} \left\| \nabla F_l(w^*) \right\| = \frac{2\sqrt{2L}}{\mu} \Tilde{p}^l\sqrt{\Gamma_l}
    \end{split}
    \end{equation*}
\end{itemize}

\subsection{Proof of Corollary \ref{cor:compare}}
\subsubsection{Scheme A}
In Scheme A, we only consider devices whose $s_\tau^k = E$. Let $q_\tau^k$ be an indicator denoting if client $k$ is complete in round $\tau$. Thus, $K_\tau = \sum_{k=1}^N q_\tau^k$.

\textbf{Homogeneous participation}. Obviously $q_\tau^k$'s are homogeneous when $s_\tau^k$'s are homogeneous. Thus, $\mathbb{E}[q_\tau^k] = q_\tau$, where $q_\tau = \mathbb{P}(s_\tau = E)$. We then have $\mathbb{P}(K_\tau = 0) = (1-q_\tau)^N$. When choosing $p_\tau^k = \frac{Np^k}{K_\tau}q_\tau^k$, $\theta = N$. Note that by the definition of $q_\tau^k$, we have $q_\tau^k s_\tau^k = Eq_\tau^k$, so $\mathbb{E}[p_\tau^ks_\tau^k] = E\mathbb{E}[p_\tau^k]$. Similarly, we can replace all $s_\tau^k$ terms with $E$. Next we calculate $\mathbb{E}[p_\tau^k]$:

\begin{equation}
\begin{split}
        &\mathbb{E}_q[p_\tau^k|K_\tau \neq 0] = Np^k\mathbb{E}_q\Big[\frac{q_\tau^k}{\sum_{i=1}^N q_\tau^i}\Big | K_\tau \neq 0]\\
        =&Np^k\sum_{i=0}^{N-1}\frac{1}{1+i} {N-1 \choose i} \frac{(q_\tau)^i(1-q_\tau)^{N-1-i}q_\tau}{1-(1-q_\tau)^N}\\
        =&Np^k\sum_{i=1}^{N}\frac{1}{i}{N-1 \choose i-1}\frac{(q_\tau)^i(1-q_\tau)^{N-i}}{1-(1-q_\tau)^N} = Np^k\sum_{i=1}^{N}\frac{1}{i}\frac{(N-1)!}{(i-1)!(N-i)!}\frac{(q_\tau)^i(1-q_\tau)^{N-i}}{1-(1-q_\tau)^N}\\
        =&p^k\sum_{i=1}^N {N \choose i}\frac{(q_\tau)^i(1-q_\tau)^{N-i}}{1-(1-q_\tau)^N} = p^k \frac{1-{N \choose 0}(q_\tau)^0(1-q_\tau)^N}{1-(1-q_\tau)^N} = p^k
\end{split}
\end{equation}

Similarly,
\begin{equation}
\begin{split}
        &\mathbb{E}_q[(p_\tau^k)^2|K_\tau \neq 0] = (Np^k)^2\mathbb{E}_q\Big[\frac{q_\tau^k}{(\sum_{i=1}^N q_\tau^i)^2}\Big | K_\tau \neq 0]\\
        =&(Np^k)^2\sum_{i=0}^{N-1}\frac{1}{(1+i)^2} {N-1 \choose i} \frac{(q_\tau)^i(1-q_\tau)^{N-1-i}q_\tau}{1-(1-q_\tau)^N}\\
        =&(Np^k)^2\sum_{i=1}^{N}\frac{1}{i^2}{N-1 \choose i-1}\frac{(q_\tau)^i(1-q_\tau)^{N-i}}{1-(1-q_\tau)^N} = N(p^k)^2\sum_{i=1}^{N}\frac{1}{i}{N \choose i}\frac{(q_\tau)^i(1-q_\tau)^{N-i}}{1-(1-q_\tau)^N}\\
        =&N(p^k)^2\mathbb{E}\Big[\frac{1}{K_\tau}|K_\tau \neq 0\Big]
\end{split}
\end{equation}

It is possible that $\sum_{k=1}^N p_\tau^k > 2$, so we need to calculate $\mathbb{E}[p_\tau^kp_\tau^l|K_\tau \neq 0]$
\begin{equation}
\begin{split}
    &\mathbb{E}_q[p_\tau^kp_\tau^l|K_\tau \neq 0] = N^2p^kp^l\mathbb{E}\Big[\frac{q_\tau^kq_\tau^l}{(\sum_{i=1}^Nq_\tau^i)^2}|K_\tau \neq 0\Big]\\
    =&N^2p^kp^l \sum_{i=0}^{N-2}\frac{1}{(2+i)^2}{N-2 \choose i}\frac{(q_\tau)^i(1-q_\tau)^{N-2-i}(q_\tau)^2}{1-(1-q_\tau)^N}\\
    =& \frac{N}{N-1}p^kp^l\sum_{i=2}^N\frac{i-1}{i}{N \choose i}\frac{(q_\tau)^i(1-q_\tau)^{N-i}}{1-(1-q_\tau)^N} = \frac{N}{N-1}p^kp^l\mathbb{E}\Big[1-\frac{1}{K_\tau}|K_\tau \neq 0\Big]
\end{split}
\end{equation}

For all $k$ and $\tau$, $\mathbb{E}[p_\tau^ks_\tau^k|K_\tau \neq 0] = Ep^k$, thus $z_\tau = 0, M_\tau = 0$ for all $k,\tau$.

Therefore,
$\mathbb{E}[{B}] = O(N^2\mathbb{E}[\frac{1}{K_\tau}|K_\tau \neq 0]+\sum_{k=1}^N(p^k\sigma_k)^2 + \Gamma), \gamma = O(N)$, hence $V = O(N^2\mathbb{E}[\frac{1}{K_\tau}|K_\tau \neq 0]+\sum_{k=1}^N(p^k\sigma_k)^2 + \Gamma)$. Plug them into Theorem \ref{tm:convergence0}, we can get an asymptotic rate of $O\left(\frac{\mathbb{E}[\frac{N^2}{K_\tau}]+\Bar{\sigma}^2_N + \Gamma}{\tau}\right)$.

\textbf{Heterogeneous Participation}. When $s_\tau^k$'s (i.e., $q_\tau^k$'s) are heterogeneous, generally $\mathbb{E}[p_\tau^k] \neq p^k$, furthermore, we may have $z_\tau = 1$ for all $\tau$. To see this, consider an example where a device $k_0$ has $q_\tau^{k_0} = 1$, i.e. $\mathbb{P}(s_\tau^k = E) = 1$, whereas all the rest devices have $\mathbb{E}[q_\tau^k] = q_\tau$, then we can show that

\begin{equation}
    \mathbb{E}_q[p_\tau^{k_0}|K_\tau \neq 0] = \mathbb{E}_q[p_\tau^{k_0}] = Np^{k_0}\mathbb{E}_q\Big[\frac{q_\tau^{k_0}}{(\sum_{i=1}^N q_\tau^i)^2}\Big] = p^{k_0}\frac{1-(1-q_\tau)^N}{q_\tau}
\end{equation}

and for $k \neq k_0$

\begin{equation}
\begin{split}
    &\mathbb{E}_q[p_\tau^{k}|K_\tau \neq 0] = \mathbb{E}_q[p_\tau^{k}] = Np^{k}\mathbb{E}_q\Big[\frac{q_\tau^{k}}{(\sum_{i=1}^N q_\tau^i)^2}\Big] \\
    =&\frac{p^k}{(N-1)q_\tau}\sum_{i=2}^N(i-1){N \choose i}(q_\tau)^i(1-q_\tau)^{N-i}\\
    =&\frac{p^k}{(N-1)q_\tau}\Big(Nq_\tau - Nq_\tau(1-q_\tau)^{N-1} - (1-(1-q_\tau)^N - Nq_\tau(1-q_\tau)^{N-1})\Big)\\
    =&p^k\frac{Nq_\tau+(1-q_\tau)^N-1}{(N-1)q_\tau}
\end{split}
\end{equation}

Thus, different $k$ will have different ratio of $\mathbb{E}[p_\tau^ks_\tau^k/p^k]=E\mathbb{E}[p_\tau^k/p^k]$, which indicates $z_\tau = 1$. Since this is true for all $\tau$, we have $M_\tau = \tau$. Thus according to Theorem \ref{tm:convergence0}, the learning will not converge to the global optimal, and the remainder loss is bounded by $D/E$.

\subsubsection{Scheme B}
In Scheme B, $p_\tau^k = p^k$ is a fixed number, so we only need to take expectation over $s_\tau^k$, and $c_p = 1$. Since $\sum_{k=1}^N p^k = 1 < 2$, we can bound $\mathbb{E}[(\sum_{k=1}^Np_\tau^k-2)_+(\sum_{k=1}^Np_\tau^ks_\tau^k)] < 0$.

\textbf{Homogeneous Participation}. When $s_\tau^k$'s are homogeneous, i.e. $s_\tau^k \overset{iid}{\sim} s_\tau$, then $\mathbb{E}[p_\tau^ks_\tau^k]/p^k = \mathbb{E}[s_\tau]$. This is the same for all $k$, thus $z_\tau = 0$ and $M_\tau = 0$. Moreover, we have $\mathbb{E}[B] = O(\mathbb{E}[s_\tau](\Bar{\sigma}^2_N + \Gamma))$, $\gamma = O(1/\mathbb{E}[s_\tau])$, $V = O\left(\left(\Bar{\sigma}^2_N + \Gamma\right)\frac{1}{\mathbb{E}[s_\tau]}\right)$, which yields an asymptotic convergence rate of $O\left(\frac{ \Bar{\sigma}^2_N + \Gamma}{\tau\mathbb{E}[s_\tau]}\right)$.

\textbf{Heterogeneous Participation}.
When $s_\tau^k$'s are heterogeneous, $\mathbb{E}[p_\tau^ks_\tau^k]/p^k = \mathbb{E}[s_\tau^k]$ varies with $k$. Thus, $z_\tau = 1$ and $M_\tau = \tau$. Therefore, the algorithm will not converge to the global optimum according to Theorem \ref{tm:convergence0}.

\subsubsection{Scheme C}
In Scheme C, $p_\tau^k = \frac{Ep^k}{s_\tau^k}$, so $\theta = E$. It is possible that $\sum_{k=1}^K p_\tau^k > 2$, so we need to calculate $\mathbb{E}\Big[(\sum_{k=1}^Np_\tau^k)(\sum_{k=1}^Np_\tau^ks_\tau^k)\Big]$.

\textbf{Homogeneous Participation}.
When $s_\tau^k$'s are homogeneous, $\mathbb{E}[p_\tau^ks_\tau^k]/p^k = E$ for all $k$. Thus, $z_\tau = 0, M_\tau = 0$.

Moreover, we have
\begin{equation}
    \mathbb{E}[\sum_{k=1}^N p_\tau^k] = E\mathbb{E}\Big[\frac{1}{s_\tau}\Big]
\end{equation}
\begin{equation}
    \mathbb{E}[\sum_{k=1}^N (p_\tau^k)^2] = \Big(E\mathbb{E}\Big[\frac{1}{s_\tau}\Big]\Big)^2\sum_{k=1}^N (p^k)^2
\end{equation}
\begin{equation}
    \mathbb{E}[\sum_{k=1}^N (p_\tau^k)^2s_\tau^k] = E^2\mathbb{E}\Big[\frac{1}{s_\tau}\Big] \sum_{k=1}^N (p^k)^2
\end{equation}

\begin{equation}
    \mathbb{E}\Big[(\sum_{k=1}^Np_\tau^k)(\sum_{k=1}^Np_\tau^ks_\tau^k)\Big] = E^2\mathbb{E}\Big[\frac{1}{s_\tau}\Big]
\end{equation}

Therefore, we have $\mathbb{E}[B] = O\left(\mathbb{E}\left[\frac{1}{s_\tau}\right](\Bar{\sigma}_N + \Gamma) \right) = V$, which yields a convergence rate of $O\left(\frac{\Bar{\sigma}^2_N+ \Gamma}{\tau(\mathbb{E}\left[1/s_\tau\right])^{-1}}\right)$.

\textbf{Heterogeneous Participation}.
Even when $s_\tau^k$'s are heterogeneous, we still have $\mathbb{E}[p_\tau^ks_\tau^k]/p^k = E$ for active all $k$. Thus, $z_\tau = 0$ only if $I_\tau = 1$. Thus, $M_\tau = \sum_{t=0}^{\tau-1} I_t$. Moreover,
\begin{equation}
    \mathbb{E}[\sum_{k=1}^N p_\tau^k] = E\sum_{k=1}^Np^k\mathbb{E}\Big[\frac{1}{s_\tau^k}\Big]
\end{equation}
\begin{equation}
    \mathbb{E}[\sum_{k=1}^N (p_\tau^k)^2] = E^2\sum_{k=1}^N \Big(p^k\mathbb{E}\Big[\frac{1}{s_\tau^k}\Big]\Big)^2
\end{equation}
\begin{equation}
    \mathbb{E}[\sum_{k=1}^N (p_\tau^k)^2s_\tau^k] = E^2\sum_{k=1}^N (p^k)^2\mathbb{E}\Big[\frac{1}{s_\tau^k}\Big] 
\end{equation}

\begin{equation}
    \mathbb{E}\Big[(\sum_{k=1}^Np_\tau^k)(\sum_{k=1}^Np_\tau^ks_\tau^k)\Big] = E^2\sum_{k=1}^Np^k\mathbb{E}\Big[\frac{1}{s_\tau^k}\Big]
\end{equation}

Thus, $\mathbb{E}[B] = O\left(\sum_{k=1}^N (p^k\sigma_k)^2\mathbb{E}\left[\frac{1}{s_\tau^k}\right] + \Gamma\right) = V$,  and the convergence rate is $O\left( \frac{\sum\limits_{t=0}^{\tau - 1}I_t D + \sum\limits_{k}^N(p^k\sigma_k)^2\mathbb{E}\left[\frac{1}{s_\tau^k}\right] + \Gamma}{\tau}\right)$

\subsection{Proof of Corollary \ref{coro:fast_reboot}}

We first introduce the following lemma:

\begin{lemma} \label{lm:expand_Fl}
Suppose device $l$ arrives, then for any $w$, we have

\begin{equation}
    F_l(w) = \frac{1}{\Tilde{p}^l} \left(\Tilde{F}(w) - \frac{n}{\Tilde{n}}F(w)\right)
\end{equation}
\end{lemma}

\begin{proof} We expand the right hand side expression and show it equals $F_l(w)$:
\begin{equation}
\begin{split}
    \frac{1}{\Tilde{p}^l} \left(\Tilde{F}(w) - \frac{n}{\Tilde{n}}F(w)\right) &= \frac{1}{\Tilde{p}^l} \left(\sum_{k=1}^{N}\Tilde{p}^kF_k(w) + \Tilde{p}^lF_l(w) - \sum_{k=1}^N\frac{n}{\Tilde{n}}p^kF_k(w) \right) \\
    &= \frac{1}{\Tilde{p}^l} \left(\sum_{k=1}^{N}\Tilde{p}^kF_k(w) + \Tilde{p}^lF_l(w) - \sum_{k=1}^N\Tilde{p}^kF_k(w) \right) = F_l(w)
\end{split}
\end{equation}
\end{proof}

Next we investigate the effect of applying additional update from $l$. Suppose the current global weight is $w_{\tau E}^{\mathcal{G}}=w$, and assume we perform full batch gradient for the additional update. After the update, it becomes

\begin{equation}
    w' = w - \eta_\tau \delta^l \nabla F_l(w)
\end{equation}

We are interested in the distance between $w'$ and the new global optimum $\Tilde{w}^*$:
\begin{equation}
\begin{split}
     \|w' - \Tilde{w}^* \|^2 &= \|w - \eta_\tau \delta^l \nabla F_l(w) - \Tilde{w}^*\|^2\\
     &= \|w - \Tilde{w}^* \|^2 \underbrace{-2  \eta_\tau \delta^l\langle w - \Tilde{w}^*,  \nabla F_l(w)\rangle + \left(\eta_\tau \delta^l\right)^2\|\nabla F_l(w) \|^2}_{A(w,\delta^l)}
\end{split}
\end{equation}

Obviously, the additional update helps fast-reboot if $A(w,\delta^l) < 0$.

Applying Lemma \ref{lm:expand_Fl} we can get
\begin{equation}
    A(w,\delta^l) = - 2\frac{\eta_\tau \delta^l}{\Tilde{p}^l}\langle w - \Tilde{w}^*,  \nabla\Tilde{F}(w) - \frac{n}{\Tilde{n}} \nabla F(w)\rangle + \left(\eta_\tau \delta^l\right)^2\|\nabla F_l(w) \|^2
\end{equation}

Write $b = w - w^*$, and use the mean value theorem we have

\begin{equation}
\begin{split}
    -\langle w - \Tilde{w}^*, \nabla F_l(w) \rangle &= -\langle b + w^* - \Tilde{w}^*, \nabla F_l(w^*) + \nabla^2F_l(\xi)b \rangle \\
    &= -\langle w^* - \Tilde{w}^*, \nabla F_l(w^*) \rangle - \langle w^* - \Tilde{w}^*, \nabla^2F_l(\xi)b \rangle - \langle b, \nabla F_l(w) \rangle \\
    &\leq -\langle w^* - \Tilde{w}^*, \nabla F_l(w^*) \rangle + \|w^*-\Tilde{w}^* \| \|\nabla^2F_l(\xi) \|_2 \|b\| + \|\nabla F_l(w) \| \|b\| \\
    &\leq -\langle w^* - \Tilde{w}^*, \nabla F_l(w^*) \rangle + (\|w^*-\Tilde{w}^* \| + 1)W\|b\| \\
    &\leq  -\frac{1}{\Tilde{p}^l}\langle w^* - \Tilde{w}^*, \nabla\Tilde{F}(w^*) - \frac{n}{\Tilde{n}} \underbrace{\nabla F(w^*) }_{=0}\rangle + \left(\frac{2\sqrt{2L}}{\mu} \Tilde{p}^l\sqrt{\Gamma_l} + 1\right)W \|b\| \\
    &\leq -\frac{1}{\Tilde{p}^l} \left(\Tilde{F}(w^*) - \Tilde{F}(\Tilde{w}^*)\right) + \left(\frac{2\sqrt{2L}}{\mu} \Tilde{p}^l\sqrt{\Gamma_l} + 1\right)W \|b\|
\end{split}
\end{equation}

Therefore,

\begin{equation}
    A(w,\delta^l) \leq 2\frac{\eta_\tau \delta^l}{\Tilde{p}^l} \left( \left(\frac{2\sqrt{2L}}{\mu} \Tilde{p}^l\sqrt{\Gamma_l} + 1\right)W \|b\| - \left(\Tilde{F}(w^*) - \Tilde{F}(\Tilde{w}^*)\right)\right) + (\eta_\tau \delta^l)^2W^2
\end{equation}

For $\delta^l > 0$, the right hand side can be negative if and only if $\|b\| <\frac{\Tilde{F}(w^*)-\Tilde{F}(\Tilde{w}^*)}{\left(\frac{2\sqrt{2L}}{\mu} \Tilde{p}^l\sqrt{\Gamma_l} + 1\right)\Tilde{p}^lW}$.

\subsection{Proof of Corollary \ref{cor:departure}}
The loss bound without objective shift is $f_{\textrm{0}}(\tau) = \frac{(\tau - \tau_0)D + V}{\tau E + \gamma}$, and the bound with shift is $f_1(\tau) = \frac{\frac{V}{\tau_0E + \gamma} + \Gamma_l}{(\tau - \tau_0)E + \gamma}$. Note that $f_0(\tau)$ is a monotonic function. When it is increasing, we just need $f_0(\tau_0) = f_1(\tau)$, which yields

\begin{equation}
    \tau - \tau_0 = \frac{1-\gamma}{E} + \frac{\Gamma_l(\tau_0 E + \gamma)}{E V} = O\left(\frac{\Gamma_l \tau_0}{V}\right)
\end{equation}

Now we consider monotonically decreasing $f_0(\tau)$, which is more commonly observed in experiments. Let $C_1 = DE, C_2 = \gamma D + VE - E\Gamma_l$, $C_3 = V(\gamma-1)$, the only possible root for the quadratic equation $f_0(\tau) = f_1(\tau)$ is

\begin{equation}
\begin{split}
        \tau - \tau_0 &= \frac{EV}{\tau_0 E + \gamma} - C_2 + \sqrt{4C_1\Gamma_l (\tau_0 E + \gamma) + \left(\frac{EV}{\tau_0E + \gamma}\right)^2 - \frac{2C_2EV}{\tau_0 E + \gamma} + \left(C_2^2 - 4C_1C_3\right)} \\
        &= O(\sqrt{\tau_0 \Gamma_l })
\end{split}
\end{equation}
\or\fi

\end{document}